\documentclass[5p, times]{elsarticle}

\usepackage{amssymb}
\usepackage{lscape}
\usepackage{longtable}
\usepackage{amsmath}
\usepackage{tabularx}
\DeclareUnicodeCharacter{221A}{\sqrt{}}
\usepackage{caption}
\usepackage{float}
\usepackage{xcolor}
\usepackage{bbding}
\usepackage{graphicx}
\usepackage{array}
\usepackage{tabulary}
\newcolumntype{K}[1]{>{\centering\arraybackslash}p{#1}}

\journal{Computer Science Review}

\begin{document}

\begin{frontmatter}



\title{Attention-Based Transformer Models for Image Captioning Across Languages: An In-depth Survey and Evaluation}


\author[inst1]{Israa A. Albadarneh}

\affiliation[inst1]{organization={The University of Jordan},
            city={Amman},
            postcode={11941}, 
            country={Jordan}}

\author[inst1,inst2]{Bassam H. Hammo}
\author[inst1]{Omar S. Al-Kadi*}
\ead{o.alkadi@ju.edu.jo}
\cortext[cor1]{Corresponding author}

\affiliation[inst2]{organization={Princess Sumaya University for Technology},
            city={Amman},
            postcode={11941}, 
            country={Jordan}}

\begin{abstract}
Image captioning involves generating textual descriptions from input images, bridging the gap between computer vision and natural language processing. Recent advancements in transformer-based models have significantly improved caption generation by leveraging attention mechanisms for better scene understanding. While various surveys have explored deep learning-based approaches for image captioning, few have comprehensively analyzed attention-based transformer models across multiple languages. This survey reviews attention-based image captioning models, categorizing them into transformer-based, deep learning-based, and hybrid approaches. It explores benchmark datasets, discusses evaluation metrics such as BLEU, METEOR, CIDEr, and ROUGE, and highlights challenges in multilingual captioning. Additionally, this paper identifies key limitations in current models, including semantic inconsistencies, data scarcity in non-English languages, and limitations in reasoning ability. Finally, we outline future research directions, such as multimodal learning, real-time applications in AI-powered assistants, healthcare, and forensic analysis. This survey serves as a comprehensive reference for researchers aiming to advance the field of attention-based image captioning.

\end{abstract}




\end{frontmatter}



\section{Introduction}

\begin{figure*}[ht!]
\centering

\includegraphics[width=13cm]{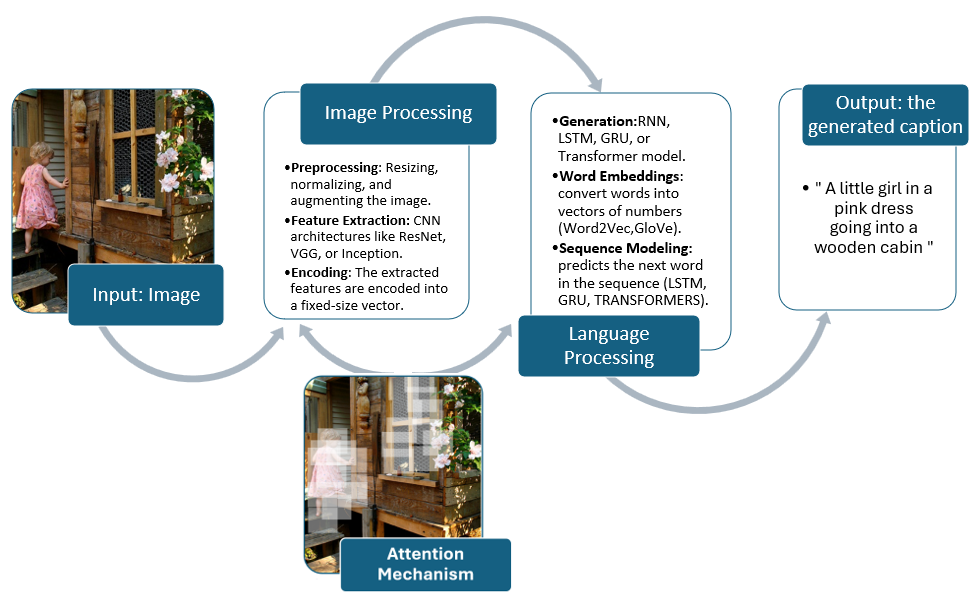}

\caption{A general architecture for image captioning models.}
\label{General_ICM}
\end{figure*}

Image captioning involves generating an image description, which includes identifying important objects and their relationships and creating syntactically and semantically correct sentences. This task requires collaboration between the computer vision (CV) and natural language processing (NLP) research communities \cite{bai2018survey} \cite{hossain2019comprehensive}. The large volume of unannotated images on the Internet has driven the automated image captioning process \cite{tien2020image}. Furthermore, advances in deep learning models have significantly improved computer vision and natural language processing capabilities \cite{cheikh2020active} \cite{yu2019multimodal}. Fig.\ref{General_ICM} shows the general architecture of the image captioning model.

The process of image captioning starts with an input image. The next step is image processing, which involves resizing, normalizing, and augmenting the image. This is followed by feature extraction using CNN architectures like ResNet or Inception, which encode the features into a fixed-size vector.
In the language processing stage, models like RNN, LSTM, GRU, or Transformer convert words into vectors and predict the next word in the sequence. The attention mechanism selectively focuses on different parts of the image to enhance the captioning process.
Finally, in the output stage, a descriptive caption is generated. For example, "A little girl in a pink dress going into a wooden cabin." This architecture effectively combines computer vision and natural language processing to automatically generate descriptive text for images.

The field of computer vision has made significant progress in tasks such as object recognition, image segmentation, image classification, and scene recognition. However, generating a natural language description of an image using a computer is generally a complex task \cite{biswas2020towards}.

Image captioning combines research from the computer vision and natural language processing communities \cite{bai2018survey}. A captioning model aims to present a scene and text, a task that comes naturally to the human brain. It is common for humans to interpret information quickly from an image at one glance. However, challenges such as parallax errors make image captioning a complex problem that is not fully resolved. This error can make it difficult for the human eye and computer vision systems to detect objects at certain angles where their appearance changes, making them hard to recognize. In addition, objects of the same class might have various shapes and appearances from different angles, further complicating the task. Overlapping objects and scene clutter also pose challenges for accurate object detection \cite{10.1145/3617592}.

Image captioning approaches have three main categories: template-based, retrieval-based, and deep learning-based. Temp-late-based techniques use predefined templates with a set number of blank slots to generate captions. These approaches first recognize different objects, characteristics, and actions and then fill in the blank spaces in the templates. Although this approach can provide grammatically correct captions and relevant descriptions, the coverage, inventiveness, and complexity of the generated sentences coverage, inventiveness, and complexity are limited.
In retrieval-based systems, captions are retrieved from a set of existing captions. This approach has the advantage of generating generic and syntactically correct captions. However, it may not produce semantically accurate captions specific to the image. Both template-based and retrieval-based approaches are not flexible enough, as they rely on existing captions in the training set or hard-coded language structures \cite{hossain2019comprehensive}.

Given the challenges of using template-based and retrieval-based approaches, a third approach based on deep learning has been introduced. This approach follows recent developments in deep neural networks, widely used in computer vision and natural language processing. Deep neural networks can provide effective solutions for visual and language modeling \cite{biswas2020towards}. As a result, they have been used to improve existing systems and create many innovative approaches \cite{bai2018survey}. 
Before significant advances in deep learning methods, image captioning was done mainly using traditional machine learning-based techniques, which included feature extraction methods like Scale-Invariant Feature Transform (SIFT), Local Binary Patterns (LBP), and Histogram of Oriented Gradients (HOG) \cite{al2008combined, al2011supervised}. A classifier was then used to classify the items after feature extraction. However, traditional approaches are less preferred than deep learning-based methods, which automatically learn features due to the complexity of feature extraction from massive amounts of data \cite{10.1145/3617592}. Numerous recent publications have focused on applying deep machine learning to caption images \cite{hossain2019comprehensive}.

Deep learning algorithms are effective at handling the complexities of image captioning. This survey offers a comprehensive analysis of transformer models and attention mechanisms in image captioning, providing a current overview of the relevant literature. The discussion is structured around the following research questions.

\begin{enumerate}[a)]
    \item \textbf{RQ1:} What are the recent advancements and current status of image captioning models, particularly attention-based and transformer-based models?

    \item \textbf{RQ2:} How have transformer-based models advanced image captioning across different languages, and what challenges and limitations do these models face?

    \item \textbf{RQ3:} How is image captioning applied across various domains, and what key evaluation metrics are used to assess the quality of generated captions?

    \item \textbf{RQ4:} What are the key conclusions from recent surveys on image captioning, and how do they guide future research directions?
\end{enumerate}


\subsection {Application of image captioning}
Research in image captioning has gained significant importance across various fields. With its broad applicability and growing technological advancements, image captioning drives innovation across numerous domains in multiple real-world applications, offering significant benefits to industry and society.

In the field of medical imaging, for instance, surgeons can apply image captioning to monitor therapy progress preoperatively, intraoperatively, and postoperatively \cite{ayesha2021automatic}. In education, researchers are exploring e-learning systems that integrate image captioning to enhance web-based learning experiences \cite{chendake2021learning}\cite{ogura2018effectiveness}. For visually impaired individuals, transformer-based photo captioning frameworks are being developed to translate visual content into written and spoken descriptions, improving accessibility and promoting self-reliance \cite{muhammed2024transformer}.  Similarly, specialized image captioning models in smart local tourism are expected to power AI-driven platforms, providing enriched experiences for travelers and local businesses alike \cite{fudholi2021image}.

Beyond these applications, image captioning plays a crucial role in virtual assistants \cite{nivedita2021image}, image retrieval \cite{hoxha2020toward}, and information retrieval \cite{wang2020paic}, as well as enhancing user engagement on social media platforms \cite{shuster2019engaging}. Additionally, researchers are exploring its potential in emerging technologies such as automated self-driving cars \cite{fujiyoshi2019deep}, CCTV footage analysis \cite{shah2018cadp}, improving image search accuracy \cite{guinness2018caption}, and enhanced facial recognition systems \cite{huang2020caption}.


\subsection{Contribution and scope}

Several survey articles have been published on the subject of image captioning. Although these surveys provided a good overview of the literature on image captioning, they did not cover publications discussing image captioning techniques for various languages. In addition, new deep-learning studies have been published since the survey papers were written. The key contributions of this survey are: (a) providing a comprehensive review of the current state of image captioning for various languages, specifically attention-based models, (b) discussing the detailed design of transformer models with different attention mechanisms, and (c) addressing ongoing challenges and highlighting potential future directions for the field. To highlight the unique contributions of this survey, Table \ref{tab:compare} compares our study with previous ones on attention-based image captioning. Unlike prior reviews, this survey extensively covers multilingual models, a broader dataset range, and an in-depth evaluation of current challenges and future directions.

\begin{table*}[ht!]
    \centering
    \caption{Comparison of this survey with existing reviews on attention-based image captioning}
   \label{tab:compare}
   \begin{tabular}{K{1cm}K{1cm}p{4cm}K{1.5cm}K{2cm}p{1.5cm}p{1.5cm}}
   \hline
Survey & Year & Focus & Attention-Based Models & Multilingual Coverage & Datasets Reviewed & Evaluation Metrics \\ 
        	\hline
\cite{khaing2019attention} & 2019 & Comparative study of deep learning models & \Checkmark & \XSolidBrush & Limited (MS, COCO, Flicker8k) & BLEU, METEOR \\

\cite{chen2021survey} & 2021 & Transformer-based models & \Checkmark & \XSolidBrush & Large-scale datasets & CIDEr, SPICE \\
\cite{stefanini2022show} & 2022 & Vision-language models & \Checkmark & \XSolidBrush & MS COCO, Flickr30k & Multiple metrics \\
\cite{luo2022thorough} & 2022 & Review of datasets and metrics & \Checkmark & \XSolidBrush & Extensive dataset coverage & Detailed metric analysis \\

\cite{zohourianshahzadi2021neural} & 2021 & Attentive deep learning models for IC & \Checkmark & \XSolidBrush & MS-COCO & B4, METEOR, CIDER and SPICE\\
\cite{senior2024graph} & 2024 & Graph types used in 2D image understanding approaches &   \Checkmark & \XSolidBrush & A summary of common datasets &  A summary of performance metrics\\
\cite{pang2023survey} & 2023 & Medical image captioning & \Checkmark & \XSolidBrush & Common data set of medical IC &  Multiple metrics\\
This Survey & 2025 & Comprehensive review of attention-based models, multilingual coverage & \Checkmark & \Checkmark  (English, Arabic. Vietnamese, etc.) & Extensive dataset coverage & Detailed metric analysis \\

\hline
\end{tabular}
\bigskip
\end{table*}

{}
\subsection{Search criteria}

The survey included papers published between 2018 and 2024. A keyword search was conducted on Google Scholar using the following terms: image captioning, image description, image text generation, transformer-based image captioning, and attention-based image captioning. Google Scholar was chosen to avoid bias towards any specific publisher \cite{10.1145/2601248.2601268}. The survey covers articles on image captioning in multiple languages, including, but not limited to, English, Arabic, Vietnamese, Myanmar, and Indonesian.

\subsection{Survey structure}
\begin{landscape}
\begin{figure}[t]
\centering
\includegraphics[width=1.1\linewidth]{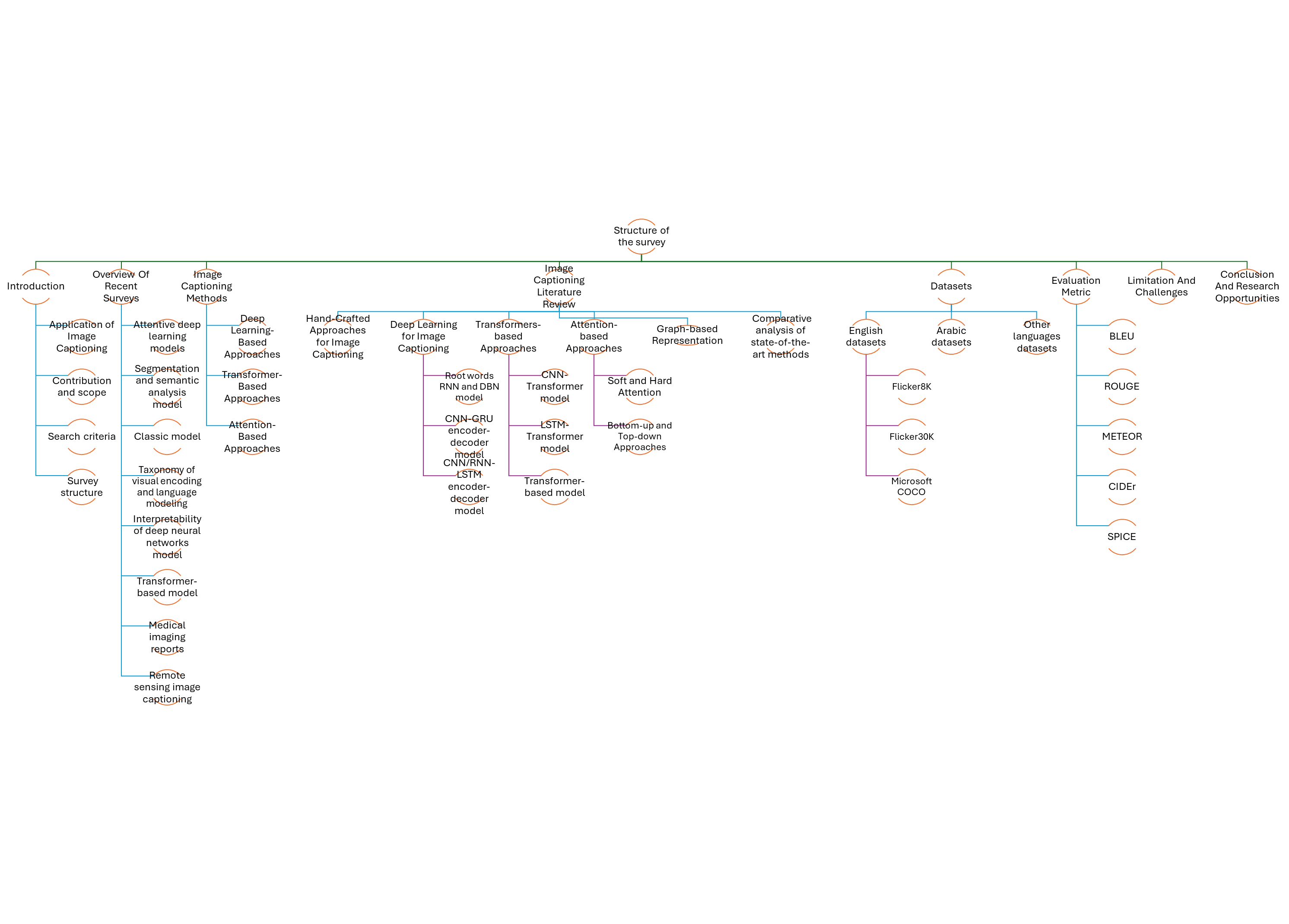}
\caption{Survey structure: Attention-based transformers for image captioning.}
\label{structure}
\end{figure}
\end{landscape}

This paper is organized as follows: Section 2 provides an overview of related surveys in image captioning, summarizes them, and discusses their foundations. Section 3 discusses the methods employed in image captioning models. Section 4 describes relevant research in image captioning, classifying them into five categories: handcrafted approaches, deep learning for image captioning, transformer-based approaches, attention-based approaches, and graph-based representation. Section 5 discusses the datasets. Section 6 introduces the evaluation metrics used to assess the quality and performance of generated captions. Limitations and challenges are summarized in Section 7, and finally, conclusions and future directions are presented in Section 8. Fig. \ref{structure} illustrates the structure of this survey.


\section{Overview of Recent Surveys}

Several recent papers have used deep learning techniques to create captions for images. This section presents a summary and analysis of relevant surveys in image captioning.
 
\subsection{Attentive deep learning models}

A literature survey by \cite{zohourianshahzadi2022neural} demonstrated that bottom-up attention models, which combine multi-head attention, yield the most significant results. \cite{khaing2019attention} proposed an attention-based deep learning model for image captioning as part of a comparative study. This research focused on attention mechanisms and identified key image areas based on the image's context, noting that attention can be beneficial in generating image captions. \cite{chen2021survey} reviewed advanced captioning techniques and classified them into attentive, semantically enhanced, transformation-based, post-editing, and vision-language pre-training (VLP).

The review by \cite{sharma2023comprehensive} examines the advancements in image captioning, tracing its evolution from traditional ML techniques to modern deep learning-based approaches. The study introduces a structured taxonomy for classifying image captioning methodologies and highlights key developments, including tem-plate-based, retrieval-based, and encoder-decoder models. Despite significant progress, the authors emphasize that further research is needed to develop more reliable and adaptable models. Similarly, \cite{sharma2024domain} explores the evolution and persistent challenges of image captioning across various application domains, such as multimodal search engines, security, remote sensing, medical imaging, and assistive technologies. The study underscores the ongoing difficulties in achieving real-time captioning in critical fields like healthcare and security and the limited availability of large, domain-specific datasets. Additionally, issues related to training and evaluation continue to pose obstacles. While significant advancements have been made, the authors stress the need for continued research to enhance the robustness and practicality of image captioning models. In another survey by \cite{sharma2022survey}, the authors focus specifically on attention-based image captioning, reviewing the major breakthroughs in this area. The paper presents a new taxonomy for classifying attention-based techniques and discusses the challenges that hinder further development. 

\subsection{Segmentation and semantic analysis models}
The application of deep learning approaches to segmentation analysis of 2D and 3D images was discussed in a study by \cite{oluwasammi2021features}. The study highlighted that while applying deep learning methods to segmentation analysis might seem straightforward for humans, it remains a challenge for computers due in part to the limited understanding of the functioning and processing mechanisms of the human brain. The study categorized supervised learning-based techniques into encoder-decoder architect-ure-based, compositional architecture-based, attention-based, semantic concept-based, stylized captions, dense image captioning, and novel object-based image captioning, as outlined by \cite{hossain2019comprehensive}.

\subsection{Classical model}

The study by \cite{stefanini2022show} aimed to identify major technical advances in architectures and training methods and to analyze various relevant state-of-the-art methodologies. This work served as a valuable resource for understanding the existing literature and outlining potential future possibilities. In addition, a systematic review of the literature (SLR) summarized advances in image captioning \cite{staniute2019systematic}. The primary objective of this research was to summarize the findings from recent papers and to describe the most popular methods and challenging problems in image captioning.

The survey \cite{sharma2023templates} explores the challenges and advancements in image captioning. The frameworks traditionally relied on a two-step pipeline, where visual features were extracted before being processed into natural language descriptions. However, with the emergence of sequential deep learning models, such as Recurrent Neural Networks (RNNs), Long Short-Term Memory (LSTM) networks, and Gated Recurrent Units (GRUs), the efficiency and accuracy of caption generation have improved significantly. The paper provides a review of the modeling architectures used, and highlights key research challenges.

\subsection{Taxonomy of visual encoding and language modeling techniques}

The purpose of the \cite{luo2022thorough} study is to present a thorough review of image captioning techniques. The researcher created a taxonomy of visual encoding and language modeling techniques, emphasizing their essential features and restrictions. A comprehensive survey was presented in \cite{hossain2019comprehensive}. This survey report provided an analysis of current deep learning-based image captioning methods, and a taxonomy of image captioning methods was provided. This study concluded that although deep learning-based image captioning techniques have made significant strides in recent years, a reliable technique still needs to be developed.

The survey by \cite{senior2024graph} examines the role of graph neural networks (GNNs) in 2D image understanding, a challenging problem in computer vision that aims to achieve human-level scene comprehension. Graphs are widely used in 2D image understanding pipelines as they effectively represent the relational structure between objects in an image. The study provides a detailed taxonomy of graph types employed in this domain, an extensive review of GNN models applied to 2D image understanding, and a forward-looking roadmap for future research. This survey covers key applications such as image captioning, visual question answering, and image retrieval, specifically focusing on approaches that exploit GNN-based architectures.

\subsection{Interpretability of deep neural networks model}

The question of the interpretability of deep neural networks, especially in image captioning methods based on classification or classification, was discussed in a survey by \cite{liu2019survey}. Due to the highly nonlinear functions and ambiguous working mechanisms, many works have aimed to explain the characteristics of `black box' models. As deep learning models are often considered black boxes, the survey conducted by \cite{choi2021component} aims to assess the impact of each module to enhance our understanding of the model. This research conducted quantitative and qualitative analyses to study the effects of five modules: the sequential module, the word embedding module, the initial seed module, the attention module, and the search module.

\subsection{Transformer-based model}

Several studies have explored the use of deep machine learning in image captioning, with a focus on transformer-based attention algorithms. However, there is a lack of thorough investigation into using transformer-based methodologies in image captioning. This gap in existing image captioning surveys has inspired this work to provide a comprehensive review of transformer-based approaches in image captioning, particularly focusing on attention-based methods.
The attention mechanism for generating image captions is an area of increasing research interest due to its consistent relevance. Initial attempts to address this issue using transformer-based approaches have shown exceptional performance. Therefore, employing transformers to improve image captioning holds great promise \cite{luo2022thorough}.

\subsection{Medical imaging reports}

The study by \cite{pang2023survey} explores prospective advancements in the automatic generation of medical imaging reports using deep learning. Inspired by image captioning techniques, deep learning algorithms have significantly improved the efficiency and accuracy of diagnostic report generation. The paper provides a review of research efforts in this domain, focusing on deep learning architectures such as hierarchical RNN-based frameworks, attention-based models, and reinforcement learning-based approaches. Additionally, it examines the applications, underlying architectures, datasets, and evaluation methods used in medical imaging report generation. The study identifies key challenges in the field and proposes future research directions to enhance clinical applications and decision-making through more advanced report-generation methods.

Similarly, \cite{THIRUNAVUKARASU2024100648} presents an analysis of transformer networks in computer vision, with a special emphasis on their applications in natural and medical image analysis. Although transformers were initially designed for natural language processing, their recent adaptation to image-based tasks has demonstrated promising results, positioning them as a viable alternative to traditional convolutional neural networks. The review highlights core principles of the transformer’s attention mechanism, which enables effective long-range feature extraction. It explores various transformer-based architectures applied to critical tasks such as image segmentation, classification, registration, and diagnosis. The paper also highlights the current limitations of transformer networks in image analysis and outlines potential research directions to enhance their effectiveness, particularly in the context of medical imaging.

Expanding on this, \cite{koteimedical} investigates the application of Vision Transformers (ViTs) in the medical domain. Initially inspired by the success of Transformer networks in language processing, ViTs have emerged as a powerful alternative to CNNs for computer vision tasks. These models and their variants excel at capturing long-range dependencies and spatial correlations, offering substantial benefits for medical image analysis tasks, including classification, segmentation, registration, detection, and radiological report generation. The paper specifically discusses the role of transformers in medical image captioning and disease diagnosis, providing insights into commonly used medical imaging modalities in clinical practice. Furthermore, it reviews the self-attention mechanism in vision transformers as applied to disease diagnosis and automated report generation. The study concludes by identifying existing challenges in the field and suggesting potential future research directions to enhance the efficiency of AI-driven applications in healthcare.

\subsection{Remote sensing image captioning}

The study by \cite{zhao2021systematic} explores the emerging field of remote sensing image captioning, which focuses on automatically generating textual descriptions for images captured by satellites, aircraft, and drones. As an interdisciplinary task integrating computer vision and natural language processing, remote sensing image captioning has garnered significant research interest in recent years. The paper analyzes relevant articles, summarizing key technical approaches, datasets, evaluation metrics, and experimental findings from state-of-the-art methods. Additionally, it examines the field's strengths, limitations, and ongoing challenges while proposing valuable directions for future research. Similarly, \cite{lu2017exploring} investigates the challenge of generating precise and adaptable textual descriptions for remote sensing images. While significant progress has been made in related tasks such as object detection and scene classification, accurately and concisely describing remote sensing imagery remains a complex problem. To address this issue, the paper introduces a set of annotation guidelines tailored to the unique characteristics of remote-sensing images, aiming to improve captioning quality. Additionally, the authors present a large-scale aerial image dataset specifically designed for remote sensing image captioning. Extensive experiments on this dataset demonstrate that the generated English descriptions effectively capture the content of remote-sensing images.

\section{Image Captioning Methods}

This section discusses various methods in image captioning models, including deep learning-based, transformer-based, and attention-based approaches.

\subsection{Deep learning-based approaches}

The creation of image captions or descriptions can be approached in various ways. Common architectures such as CNN, RNN, and LSTM are often used to generate image captions. A convolutional neural network (CNN), an artificial intelligence (AI) network, has been utilized in many fields, including pattern recognition and natural language processing. Artificial neural networks (ANNs) are mathematical models with layers typically consisting of an input layer, an output layer, and one or more hidden layers. If $x$ represents the input and $f$ is the activation function, mathematically, a neuron can be represented as 

\begin{equation}
\label{eq1}
    z=f\left(\sum_{i=0}^n w_i x_i+\beta\right)
\end{equation}\\

\noindent where $n$ is the number of input features, $w$ is the connection weights between the input layer and the hidden layer, $\beta$ is the bias weight. 

In most deep learning models, CNN is an encoder network, while RNNs are used as language-model decoder networks. However, some image captioning models use RNN for the encoder and decoder networks. A recurrent neural network includes an LSTM (long-short-term memory) component for long-term and short-term memory. LSTM is used for sentence representation to create image captions and extract features of images and words \cite{khaing2019attention}. However, RNNs, LSTMs, and GRUs are susceptible to problems such as vanishing gradients, training difficulties, and long sequences. RNNs may not retain all information at the beginning of a long sequence. The specific operations of the LSTM-based decoder used in \cite{biswas2020towards} to generate captions are described in (\ref{eq2}), (\ref{eq3}), and (\ref{eq4}).

\begin{equation}
\label{eq2}
\left[\begin{array}{l}
i_t \\
f_t \\
o_t \\
g_t
\end{array}\right]=\left[\begin{array}{c}
\sigma \\
\sigma \\
\sigma \\
\tanh
\end{array}\right] T_{D+m+n, n}\left[\begin{array}{c}
E_{y_{(t-1)}} \\
h_{t-1} \\
\hat{z}_t
\end{array}\right]
\end{equation}

\begin{equation}
\label{eq3}
 c_t=f_t \odot c_{t-1}+i_t \odot g_t   
\end{equation}

\begin{equation}
\label{eq4}
 h_t=o_t \odot \tanh \left(c_t\right).   
\end{equation}

\begin{figure*}[ht!]
\centering
\includegraphics[width=8cm]{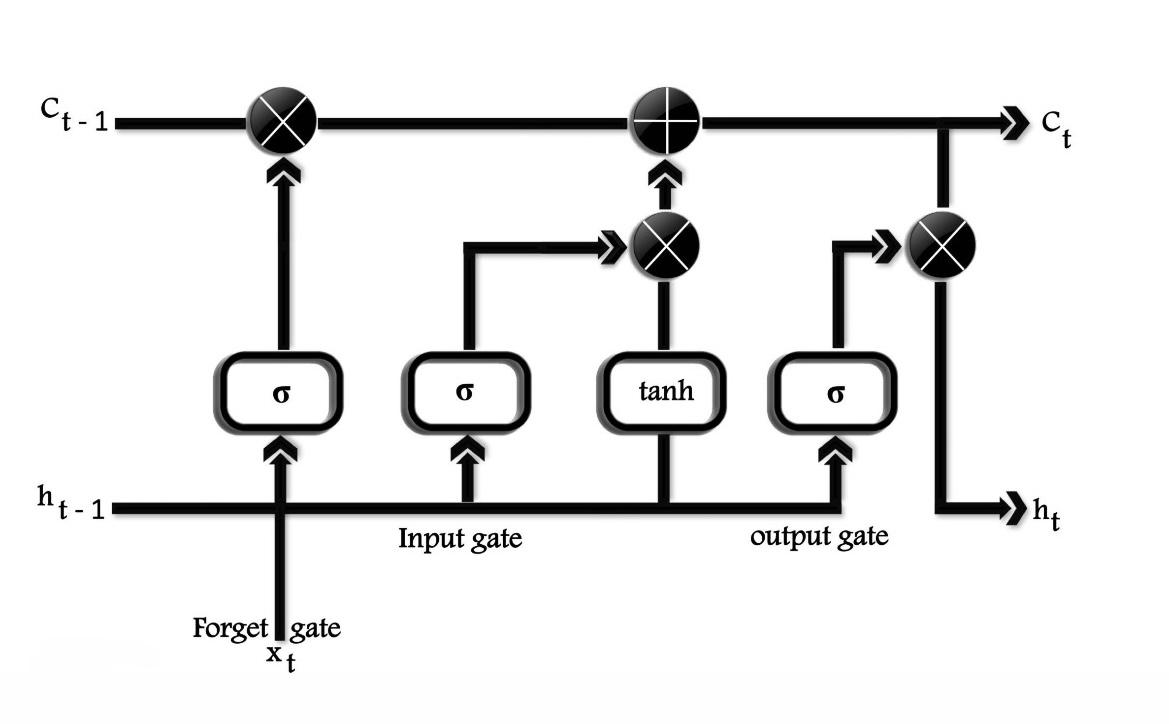}
\caption{A typical LSTM unit consisting of forget, input, and output gates.}
\label{lstm}
\end{figure*}

Input, forget, memory, output gates, and hidden state are represented by variables $i_t$, $f_t$, $c_t$, $o_t$, and $h_t$, respectively. $T$ is a mapping with the formula $f_{s, t}: \mathbb{R}^s \rightarrow \mathbb{R}^t$. As a result, $\mathbb{R}^{(D+m+n)}$ to $\mathbb{R}^n$ is mapped by $T_{D+m+n, n}$. $\hat{z} \in \mathbb{R}^D$ stands for the context vector that captures the visual data of a particular area in the input image. $E$ stands for the embedding matrix of dimension $m \times k$. The dimension of the embedding vector is indicated by the letter $m$, and the letter $n$ indicates the dimension of the hidden state LSTM. Furthermore, $\sigma$ and $\odot$ represent logistic sigmoid and element-wise multiplication, respectively. A typical LSTM unit is shown in Fig.\ref{lstm}.\\

The Long Short-Term Memory Network (LSTM) is a type of recurrent neural network (RNN) known for its superior performance. However, training LSTM networks can be challenging due to the complex addressing and overwriting mechanisms, the inherently sequential nature of the required processing, and the significant amount of storage needed during the procedure \cite{10.1145/3617592}. While LSTMs are slower at processing than CNNs, they excel at modeling dynamic temporal behavior in language, which cannot be achieved using only a language model \cite{hossain2019comprehensive}. On the other hand, global CNN features are known for their ease of use and compact representation. However, this approach also leads to excessive information compression and requires granularity, making it difficult for a captioning model to provide detailed descriptions \cite{stefanini2022show}.

\begin{figure*}[ht!]
\centering
\includegraphics[width=11cm]{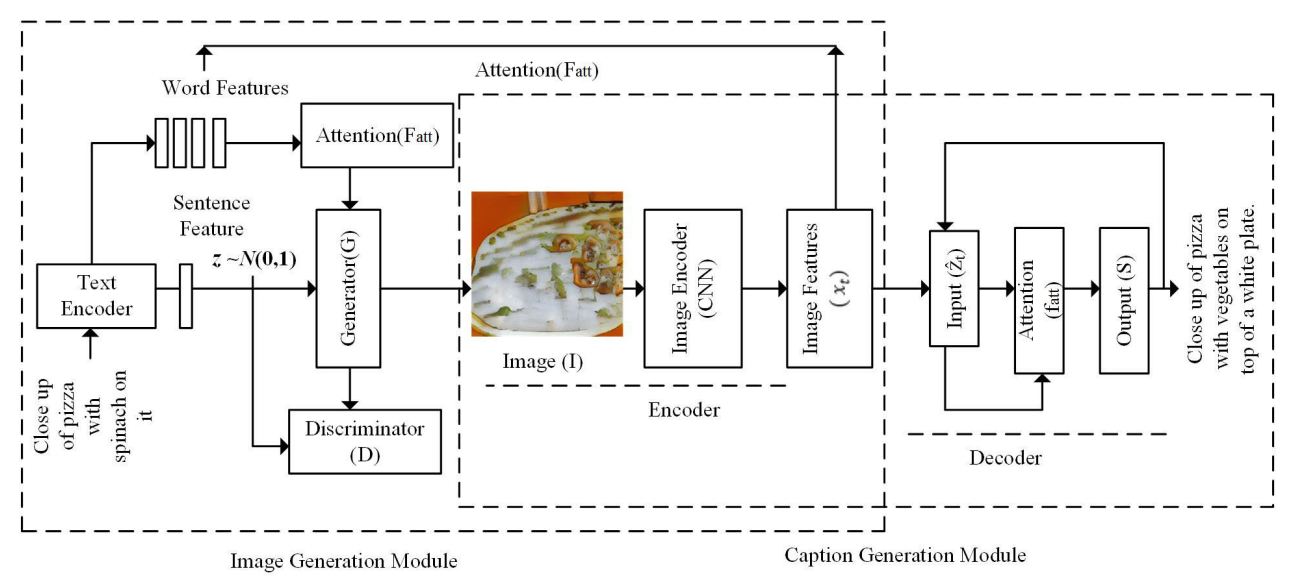}
\caption{GAN-based model for image captioning
\cite{hossain2021text}.}
\label{GAN}
\end{figure*}

In machine learning, another technique is reinforcement learning, while unsupervised learning methods include generative adversarial networks (GANs). GAN-based image captioning systems are capable of producing a variety of image descriptions. However, text processing relies on discrete numbers, making the processes non-differentiable and challenging to apply back-propagation directly. The architecture of the method presented by \cite{hossain2021text} is shown in Fig.\ref{GAN}. It uses a GAN-based model to generate artificial images from text, employing attention to focus on relevant word vectors to create various parts of the image. Subsequently, captions are produced for the image using an attention-based image captioning model. \cite{zhang2021consensus} introduced a Gated Recurrent Unit (GRU) based on the generative adversarial structure network (GASN), which consists of three parts: a consensus reasoning module, a sentence decoder with two layers of LSTM, and a grounding module to locate regions. This method provided accurate and detailed information on objects to predict words.\\

\subsection{Transformer-based approaches}

The transformer is a neural network architecture introduced in \cite{vaswani2017attention}. It excels at handling sequential text data and comprises a stack of encoder and decoder layers. Each encoder and decoder stack contains the corresponding embedding layers for their inputs and an output layer to generate the final output. The encoder includes a self-attention layer for calculating relationships between words in the sequence, a feedforward layer, and a second encoder-decoder attention layer. Residual skip connections and two LayerNorm layers surround the encoder and decoder layers. Data inputs for the encoder and decoder include the embedding and position encoding layers. The encoder stack consists of multiple encoders, each with a feedforward and multi-head attention layers. In contrast, the decoder stack includes multiple decoders, each with two feedforward layers and multi-head attention \cite{vaswani2017attention}.

Recent image captioning models leverage transformer architectures to connect informative regions in the image using attention, resulting in excellent performance. However, some previous transformer-based image captioning models have limitations because the transformer's internal architecture was originally designed for machine translation. Text sequences are inherently sequential, whereas images are two- or three-dimensional, leading to significant differences in the relative spatial relationships between regions in images compared to phrases \cite{he2020image}.

The transformer consists of two main parts: an encoder and a decoder. Multi-head attention functions as parallel heads of self-attention. Self-attention is the mechanism used by transformers to incorporate the context of other relevant words into the processing of the current word. Another component is the fully connected feedforward network, consisting of two linear transformations consistent across positions but varying parameters from layer to layer. The transformer adds a vector to each input embedding to help determine the position of each word; position embedding is a way of considering the order of words in an input sequence. The linear layer is a simple, fully connected neural network that transforms the vector produced by the decoder stack into a much larger vector known as a logit vector. SoftMax provides the probabilities. The cell with the highest probability is chosen, and the word associated with it is produced as the output \cite{vaswani2017attention}.

The transformer model addresses the limitations of RNN and LSTM by enabling more parallelization and improving translation quality. Unlike RNN or LSTM, which process sentences one word at a time, transformer models can handle complete sentences through attention-based mechanisms \cite{wolf2020transformers}. Although RNN has challenges in scaling to larger levels, attention-guided image captioning can outperform later transformer-based techniques when used with strong visual encoders. Although these methods are often smaller than transformer-based approaches, they require longer training times. The transformer-based approach resolves the issue of long-distance dependency present in RNN. Furthermore, its structure makes it easier to scale the transformer model to deeper levels following the actual design requirements \cite{luo2022thorough}.

\subsubsection{The transformer model}

The transformer network uses an encoder-decoder architecture similar to RNN but with a key distinction. Unlike RNNs, transformers can simultaneously process the entire input sentence or sequence without any time step associated with the input. Transformers consist of $N$ identical layers, each containing three sub-layers. The first layer utilizes a multi-head self-attention technique, including a mechanism to prevent the model from seeing future data, ensuring that the model only uses prior words to generate the current term. The second layer performs multi-head attention over the output of the first layer, serving as the foundation for correlating text and visual information with the attention mechanism. The third layer is a fully connected feedforward network. Following layer normalization, the transformer applies a residual connection around the three sub-layers. Unlike LSTMs, the transformer can process all words in the caption simultaneously.

Transformers do not rely on recurrence or convolution and thus need to learn the relative or absolute positions of the words in a sequence. This is achieved by employing learned weights that represent the position of a token within a sentence. The fully attentive paradigm proposed by \cite{vaswani2017attention} has significantly transformed the way language production is viewed, leading the Transformer model to become the cornerstone of many NLP innovations and the \textit{de facto} standard architecture for numerous language processing tasks.

\begin{figure*}[ht!]
\centering
\includegraphics[width=10cm]{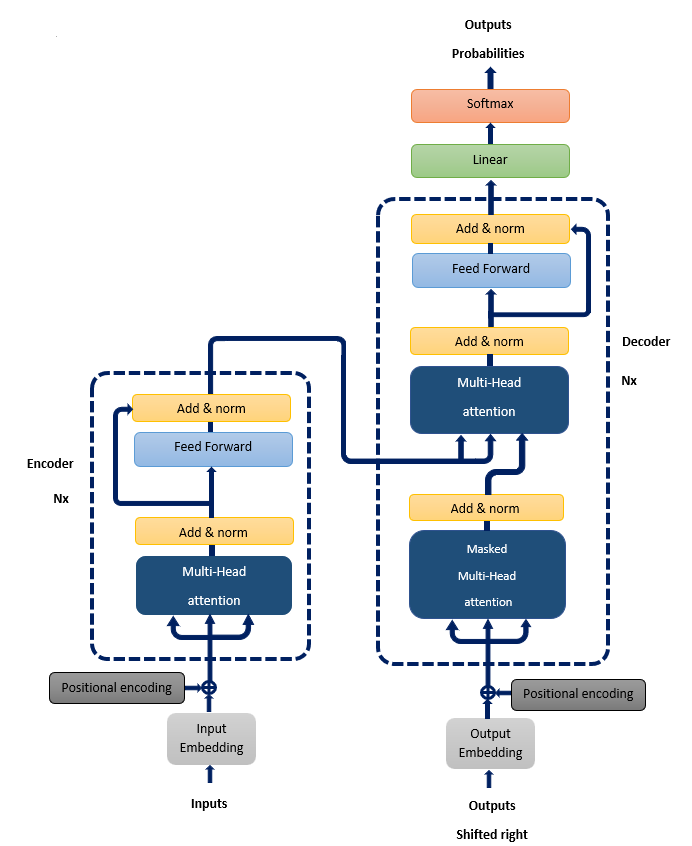}
\caption{General architecture of an encoder-decoder transformer.}
\label{Transformer_2}
\end{figure*}

The Transformer design has been utilized for image captioning, which can be considered a sequence-to-sequence task. In the conventional transformer decoder, words undergo a masked self-attention operation, followed by a cross-attention operation where words act as queries, and the output of the final encoder layer acts as keys and values, along with a final feedforward network. During training, a masking strategy is used to limit the influence of the preceding words \cite{stefanini2022show}. Both the encoder and decoder of the Transformer utilize layered self-attention and point-wise interconnected layers, as shown in the left and right halves of Fig.\ref{Transformer_2}. Self-attention, or intra-attention, focuses on the relationships between different positions in a single sequence to represent the sequence. Self-attention has been successfully applied in reading comprehension, abstractive summarization, textual entailment, and sentence representations independent of the learning task \cite{vaswani2017attention}.

The attention mechanism focuses on a subset of the details relevant to our objective instead of assessing the entire picture simultaneously. The core of the attention mechanism lies in selecting the portion of detail to concentrate on based on our goals and continually analyzing it. By calculating the similarity of word vectors, self-attention determines the degree of correlation between the current word and other words for the image captioning task. Typically, two-word vectors have smaller distance angles and greater products the closer their meanings are to each other. By normalizing the similarity, weights are generated. The attention score also referred to as the level of attention of the current word to other words, is obtained by multiplying the weights by the word vectors and summing them. The feedforward network, a one-way propagation neural network, can be classified based on the sequence in which information is received. The neurons in each layer receive the output of the neurons in the layer below and send it to the neurons in the layer above \cite{lu2023full}.

\subsubsection{Self-head attention}

The concept of self-attention involves each element in a set being related to every other element. This is achieved through a process called "self-attention," which helps to compute a more precise representation of the set using residual connections. \cite{vaswani2017attention} initially introduced this idea for language understanding and machine translation tasks. This led to the development of the Transformer architecture and its various iterations, which have been widely influential in natural language processing (NLP) and computer vision.

Self-attention can be formally explained through the scaled-dot product mechanism. It involves a multiplicative attention operator that works with three sets of vectors: a set of query vectors $Q$, a set of key vectors $K$, and a set of value vectors $V$. Each set consists of $n_k$ element-strong vectors created using linear projections of an identical input set of components.
The key and query vectors are used to compute the similarity distribution, which is then used to calculate a weighted sum of the value vectors. This process helps to capture the relationships and dependencies between different elements in the set. \cite{stefanini2022show} has further contributed to understanding self-attention and its applications.

\subsubsection{Multi-head attention}

The multi-head attention module in the transformer model utilizes the attention mechanism in parallel multiple times. This involves concatenating and linearly transforming the outputs of the attention mechanism. Multi-head attention allows for simultaneous self-attention across different sections of the input sequence \cite{wang2020transformer}, helping to capture both long-term and short-term dependencies. There are two types of attention mechanisms: soft attention and hard attention. In soft attention, weighted image features are used as input to the model instead of the raw image, enabling the model to focus on important areas and ignore less relevant ones. Soft attention uses conventional back-propagation for gradient computation and assumes that the weighted average accurately represents the focus region. On the other hand, hard attention involves sampling using the Monte Carlo approach and then averaging the results to obtain the final output. The precision of hard attention is determined by the number and quality of the samples taken \cite{10.1145/3617592}.

The drawback of attention-based approaches is the low precision in selecting the attention area, as mentioned in some articles. Most attention-based methods choose regions of the same size and shape without considering the image contents. Determining the best number of area recommendations involves a trade-off between small and huge amounts of detail. Another issue is the single-stage structure of attention-based approaches. Since most approaches have a single encoder-decoder attention structure, they cannot generate detailed captions for the images \cite {10.1145/3617592}. In a typical attention-based paradigm, an adaptive attention module learns how often to attend, while a base attention model performs a single attention step for each time step. In these methods, the characteristic of the image matches one captioning word at each time step. As the output of one attention mechanism depends directly on the outcome of another, the relationship between the attended feature and the attention inquiry is not modeled \cite {chen2021survey}.

\begin{figure*}[ht!]
\centering
\includegraphics[width=8cm]{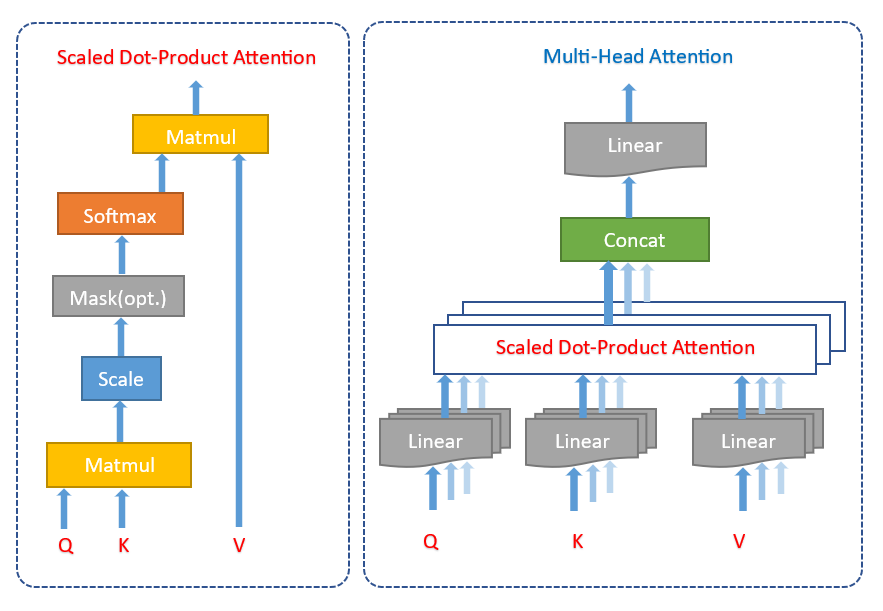}
\caption{Attention mechanism: (left) scaled dot-product attention, (right) multi-head attention.}
\label{scaled-multi}
\end{figure*}

The transformer model for neural machine translation highlighted multi-head attention effectiveness based on multiple scaled-dot attention heads. Both the encoder and decoder were constructed using multi-head attention. Currently, models prioritizing scaled-dot and multi-head attention over bottom-up characteristics and semantic information yield the best results for image captioning. Multi-head attention techniques outperform existing methods, making them the best practices when utilizing attention mechanisms for image captioning \cite{zohourianshahzadi2022neural}. 
In Fig. \ref{scaled-multi}, the scaled attention of the dot product is shown in the left block, where self-attention calculates the dot product of the query with all keys, which is then normalized using the SoftMax operator to obtain attention scores. These scores determine the weights, and each element becomes the weighted sum of all elements in the sequence. On the other hand, the right block represents multi-head attention, consisting of multiple self-attention blocks ($h = 8$ in the original Transformer model) to capture complex interactions between various items in the sequence.

\subsubsection{Add and Norm layers}

The Add and Norm layers perform two operations. The `add' step controls the flow through residual connections. The second step is `Norm,' which performs layer normalization. As a result, the output of this layer will follow (\ref{eq5}).

\begin{equation}
\label{eq5}
    \text { Add \& Norm }=\text { LayerNorm }(x+\text { Sublayer }(x))
\end{equation}

where $x$ is the input of any sublayer (MHA or feedforward), and the sublayer ($x$) is the output.

\subsubsection{ Feed Forward Network}
Each layer contains a fully connected point-wise feedforward network using ReLU activation for two linear transformations. The layer determines the weights used during training, which can be defined numerically as

\begin{equation}
\label{eq6}
\begin{gathered}
F F(x)=\operatorname{ReLU}\left(x W_1+b_1\right) W_2+b_2 \\
\operatorname{ReLU}(x)=\max (0, x)
\end{gathered}
\end{equation}

where $W_1$ and $W_2$ are network weight matrices and $b_1$ and $b_2$ are biases.

\subsubsection{Positional encoding}
The transformers incorporate positional encoding to introduce the relative or absolute positions of the tokens into the model. This helps maintain the parallel execution format of the token sequence. The positional encoding values are calculated using sine and cosine functions to represent the position and training parameters. These positional encodings are combined with language features to create embeddings that are aware of the position within the sequence.

\subsubsection{Linear and SoftMax layer}
Like in the seq2seq models, the decoder output is transformed by a fully connected linear layer to match the vocabulary size $n$, representing the expected result size. The vocabulary size of a language depends on the sentence length and the size of its vocabulary. After the transformation, a SoftMax layer is applied to the resulting matrix to create a probability distribution for each word in the output phrase over the vocabulary.

\subsubsection{Encoder and decoder stacks}
The encoder \cite{vaswani2017attention} consists of a stack of identical layers $N = 6$, each containing two sub-layers. The first sub-layer is a multi-head self-attention mechanism, and the second is a simple, positionwise, fully connected feedforward network. A residual connection around the two sublayers is used, followed by layer normalization. This allows an attention vector to capture the contextual links between words in a sentence for each word. Self-attention, a specific attention mechanism used by multi-headed attention in the encoder, enables models to connect each word in the input to other words. Similarly, the decoder comprises a stack of $N = 6$ identical layers, adding a third sublayer to each encoder layer. This additional sub-layer performs multi-head attention over the output of the encoder stack. As with the encoder, residual connections are utilized around each sub-layer, followed by layer normalization. Furthermore, the self-attention sub-layer in the decoder stack is modified to prevent positions from attending to preceding positions. This means that predictions for location $i$ can only involve known outputs at positions less than $i$ due to this masking and the offset of the output embeddings by one position. Finally, the decoder is completed by a linear layer serving as a classifier and a SoftMax to determine word probabilities.

\subsubsection{Attention function}
A set of key-value pairs, a query, and an output, all represented as vectors, can be linked using an attention function. The output is determined by calculating the weighted sum of the values, with each value's weight based on the compatibility of the query with its corresponding key.
The attention mechanism described in \cite{vaswani2017attention} is called Scaled Dot Product Attention. The input consists of queries, keys of dimension $d_k$, and values of dimension $d_v$. First, the dot product of the query with all keys is computed and then divided by $\sqrt{d_k}$. Subsequently, a SoftMax function is applied to obtain the weights of the values. The attention function is continuously computed on a group of queries gathered into a matrix $Q$. The keys and values are organized similarly into matrices $K$ and $V$. The output matrix is estimated as 

 \begin{equation}
 \label{eq7}
    \operatorname{Attention }(Q, K,V)= \operatorname{softmax}\left(\frac{Q K^T}{\sqrt{d_k}}\right) V.
 \end{equation}

In \cite{vaswani2017attention}, the usefulness of the linear projection of queries, keys, and values multiple times was demonstrated using distinct learned linear projections. The queries, keys, and values were projected to dimensions $d_q$, $d_k$, and $d_v$ rather than using a single attention function with model-dimensional keys, values, and queries. The attention function was applied to these projected versions simultaneously, resulting in $d_v$-dimensional output values. The model could use multi-head attention to data from multiple representation subspaces at different locations. Their study used eight parallel attention layers, or heads, with the formula $d_k$ = $d_v$ = $d_{model}/h$ = 64 applied to each. Despite using multiple heads, the total computing cost was comparable to that of single-head attention with full dimensionality due to the lower dimension of each head as shown in (\ref{eq8}),

\begin{equation}
\label{eq8}
    \begin{aligned}
\operatorname{MultiHead}(Q, K, V) & =\operatorname{Concat}\left(\operatorname{head}_1, \ldots, \operatorname{head}_{\mathrm{h}}\right) W^O \\
\text { where head } &=\operatorname{Attention}\left(Q W_i^Q, K W_i^K, V W_i^V\right).
    \end{aligned}
\end{equation}

\subsection{Attention-based approaches}
\begin{figure*}[ht!]
\centering
\includegraphics[width=13cm]{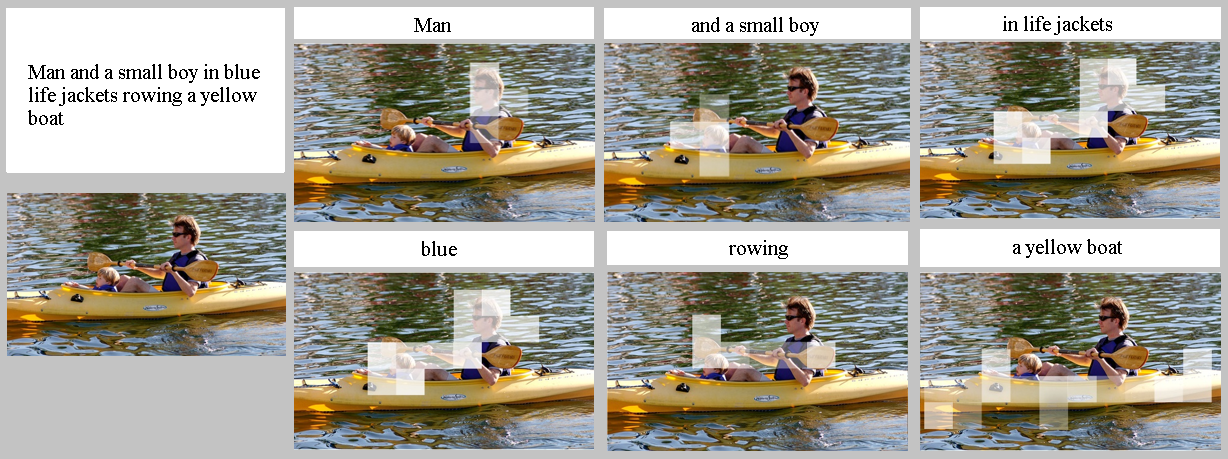}
\caption{Attention mechanism dynamics showing how the model shifts its focus across different image regions over time, aligning its attention to generate each corresponding word in the caption.}
\label{Attention}
\end{figure*}

Techniques such as CNN or RNN can generate image descriptions but cannot analyze the image over time. Moreover, these approaches do not consider the spatial elements of the image that are crucial for generating image captions. Instead, attention-based techniques are gaining popularity in deep learning, as they consider the entire context when creating captions. They can dynamically focus on different elements of the input image as the output sequences are generated. These methods commonly use CNN to gather image data and then employ a language generation phase to produce words or sentences based on the output. Each language generation step focuses on the image's prominent areas until reaching the final state. Although attention-based methods aim to identify various regions of the image when generating words or phrases for image captions, the accuracy of the attention maps produced by these methods may affect the quality of the generated captions \cite{hossain2019comprehensive}. The effectiveness of attention mechanisms in deep learning models has led researchers to emphasize their importance in image captioning \cite{zohourianshahzadi2022neural}. Fig. \ref{Attention} shows changes in attention over time as the model generates each word to reflect the relevant parts of the image \cite{xu2015show}.\\

People can focus on certain details while disregarding others when receiving information. This self-selection process is known as attention. The attention mechanism is an important development in generation-based models within the encoder-decoder architecture. It aims to improve the encoder-decoder model by imitating the human eye's focus on different areas in an image when generating descriptive words. The concept of attention originated from studying human vision in cognitive neurology, which led to the discovery of this higher brain function. The attention mechanism has diverse applications, including image categorization in visual images and various experiments in natural language processing, such as machine translation, abstract creation, text understanding, text classification, and visual captioning \cite{wang2020overview}.

Human attention patterns and visual focus on images have inspired attention-based approaches. In these mechanisms, the model is directed to pay more attention to the most important characteristics of an image, similar to how humans do. The attention mechanism guides the model on "where to look" during the training process \cite{10.1145/3617592}. It is recognized that images contain a vast amount of information, but not all features need to be explained in the captioning of images. Instead, the focus should be on the most essential content. If attention is integrated into the encoder-decoder picture captioning framework, sentence creation will be influenced by hidden states computed using the attention method. This framework incorporates an attention mechanism that allows the decoding process to concentrate on specific features of the input image at each time step to generate a description of the image \cite{bai2018survey}.

The human visual system inspires the mechanism of attention in image processing. Like our eyes do not take in every detail of an image at once, the attention mechanism also focuses on the key elements before moving on to the next. This approach is believed to enhance image captioning by eliminating irrelevant information. By mimicking the cognitive function of human vision, this mechanism can also reduce computational load and improve training accuracy \cite{liu2019survey}.

Attention mechanisms are widely used in applications such as image captioning, machine translation, speech recognition, image synthesis, and visual question-answering models \cite{yu2019multimodal} \cite{pedersoli2017areas}. Attention has been shown to connect the meaning of features, which aids in understanding how one aspect relates to another. Incorporating this into a neural network helps the model focus on the most important and relevant features while ignoring other noisy parts of the data space distribution \cite{oluwasammi2021features}.

\section{Image Captioning Literature Review}

This section categorizes image captioning models into Hand-Crafted Approaches for Image Captioning, Deep Learning for Image Captioning, and Transformer-Based Image Captioning. The state-of-the-art image captioning methods are provided in Table \ref{tab:results}. Furthermore, Table \ref{tab:Multilingual image captioning models} offers a summary of image captioning models for different languages.

\begin{table*}[htbp]
    \centering
    \caption{Overview of state-of-the-art methods in image captioning, highlighting key techniques, datasets, and performance metrics}
    \label{tab:results}
    \begin{tabular}{llcccccccc}
    \hline

    \hline
    \textbf{Reference} & \textbf{Dataset} & \textbf{B1} & \textbf{B2} & \textbf{B3} & \textbf{B4} & \textbf{CIDEr} & \textbf{METEOR} & \textbf{ROUGE} & \textbf{SPICE} \\
    
    \hline
    \\
    \cite{jindal2017deep} &   Arabic Al-Jazeera news$^{m}$  &  0.348   &    NA     &    NA     &     NA    &  NA     &   NA    &   NA    & NA \\ 
    \cite{jindal2018generating}  &   Arabic Flickr8k$^{m}$  &   \textbf{0.658}    &  \textbf{0.559}    &  \textbf{0.404}    &  \textbf{0.223}    &  NA &   0.209   &    NA   &   NA    \\ 
    \cite{al2018automatic} &  Arabic Flickr616  &  0.460    & 0.260    & 0.190    & 0.080     &     NA  &     NA  &    NA   &  NA     \\   
    \cite{mualla2018development} &  Arabic Flickr8k$^{x}$ & 0.344   & 0.154   & 0.076    & 0.035    &  NA     &  NA     &  NA    &   NA \\ 
    \cite{eljundi2020resources}   & Arabic Flickr8k & 0.330    & 0.190    & 0.100    & 0.060    &   NA    &   NA    &    NA   &    NA   \\  
    \cite{hejazi2021deep} & Arabic Flickr8k  & 0.365   & 0.214   & 0.120    & 0.066   &  NA     &    NA   &  NA & NA  \\ 
    \cite{sabri2021arabic} & Arabic Flickr8k  & 0.443   & NA   & NA    & 0.157   &  NA     &   \textbf{0.343}  &  NA & NA  \\  
    \cite{emami2022arabic} & Arabic Flickr8k & 0.391   & 0.246   & 0.151    & {0.093}   &  0.428     &   {0.317}   &  0.334 & NA  \\ 
     \cite{lasheen2022arabic} & Arabic Flickr8k  & 0.391   & 0.251   & 0.140    & 0.083   &  NA     &    NA   &  NA & NA  \\ 
    \cite{alsayed2023performance} & Arabic Flickr8k  & 0.489   & 0.317   & 0.213    & 0.145   & \textbf{0.472}     &    {0.334}   &  \textbf{0.398} & NA  \\ 
 \cite{elbedwehy2023improved} & Arabic Flickr8k & {0.598}   & {0.400}   & {0.306}    & {0.165}   & 0.469       &    0.260   &   {0.385} & NA  \\ 
\\     
\hline
\\
 
   \cite{karpathy2015deep} & English Flickr8k & 0.579          & 0.383          & 0.245        & 0.160          & NA                & NA               & NA              & NA              \\ 
   \cite{bineeshia2021image} & English Flickr8k & 0.589          & 0.335          & 0.263        & 0.148          & NA                & NA               & NA              & NA              \\

   \cite{ma2023towards} & English Flickr8k & 0.674	&NA &NA	& {0.243}	& {0.636} 	& {0.215}	& {0.448} &NA
   \\

  \cite{jiang2019modeling} & English Flickr8k & {0.690}	& {0.471}	& {0.324}	& 0.219	&{0.507}	& 0.203	& {0.502}	&NA
   \\

 \cite{karpathy2015deep} & English Flickr30k & 0.573  & 0.369  & 0.240 & 0.157 & NA   & NA  & NA  & NA            \\ 
\cite{do2020reference} & English Flickr30k & {0.695}	&0.463	&0.341	&0.232	&	0.486 & \textbf{0.302} &0.451	  &NA\\  

   \cite{ma2023towards} & English Flickr30k & 0.671	&NA &NA		& 0.233	& {0.645} & 0.204 & 0.443		 &NA
   \\ 

\cite{lu2017knowing} & English Flickr30k
 &0.677	&0.494	&0.354	&0.251 
&0.531
&0.204
&0.467
&{0.145}
\\ 
   
   \cite{kalimuthu2021fusion} & English Flickr30k & 0.647	& 0.456	& 0.320	& 0.224	& 0.467 & 0.197 & 0.449			& 0.136
   \\

   \cite{jiang2019modeling} & English Flickr30k & 0.689	& 0.468	&0.319	&0.220	&0.428 &0.191 &0.487			&NA
  \\

   \cite{abdussalam2023numcap} & English Flickr30k & 0.694	& {0.498}	& 0.355	& 0.254	& 0.469 & 0.251 & {0.538}			&NA
   \\ 


   \cite{shrimal2021attention} & English Flickr30k & 0.674
	& 0.495
	& {0.360}	& {0.260}
	& 0.520 & 0.201 & 0.470			&NA
   \\ 
  
    \cite{zhao2020cross} & English Flickr30k & 0.690 &	0.493	&0.347	&0.241 & 0.528  & 0.195 & 0.465 &NA
   \\

\cite{WANG2022117174} & English MS COCO & 0.744 & 0.567 & 0.418 &0.308 & 0.680 &  0.234 &NA &NA \\

\cite{xu2015show} &English MS COCO
&0.718 &0.504 &0.357 &0.250 &NA & 0.230 &NA & NA\\

\cite{10.1007/s11063-021-10431-y} &English MS COCO &0.748 &0.577 &0.428 &0.314 & 1.061 &0.265 &0.553 &NA\\

\cite{fei2022attention} &English MS COCO &0.822 &0.670 &0.524 &0.402 & 1.324 &0.297 & 0.595 &NA\\

\cite{Wang2022EndtoEndTB}  &English MS COCO & \textbf{\underline{0.828}} & \textbf{\underline{0.681}} & \textbf{\underline{0.536}} & \textbf{\underline{0.414}} & \textbf{\underline{1.360}} & 0.301 & \textbf{\underline{0.604}} &NA\\

\cite{yang2022reformer} &English MS COCO &0.823 &NA &NA &0.398 &1.319 & 0.297 & 0.598 & \textbf{0.230}\\

\\
\hline
\\
\cite{9971855} &   Indonesian Flickr8k Bahasa & \textbf{0.560} & \textbf{0.412} & \textbf{0.294} & \textbf{0.206} & \textbf{0.573} & \textbf{0.195} & \textbf{0.442} &NA\\
\cite{mulyanto2019automatic} & Indonesian FEEH-ID &0.500 &0.314 &0.239 &0.131 &NA &NA &NA &NA\\

\cite{10497459} & Indonesian Flickr8k   & 0.387 &0.211 & 0.087 & 0.032 &NA &NA &NA &NA\\

\cite{Nugraha2019GeneratingID} & Indonesian Flickr8k   & 0.360 &0.170 &0.060 &0.020 &NA &NA &NA &NA\\
\\
\hline
\\

\cite{pa2020automatic} &Myanmar Flickr8k

&0.641 &0.486 &0.399 &0.244 &NA &NA &NA &NA\\

\cite{10181326} &Myanmar corpus & \textbf{0.703} & \textbf{0.581} & \textbf{0.513} & \textbf{0.386}  &NA &NA &NA &NA\\
\\
\hline
\\
\cite{muhammad2022bornon} & Bengali BORNON &0.605 &0.492 & \textbf{0.412} & \textbf{0.351} &NA & \textbf{\underline{0.348}} &NA &NA\\

\cite{khan2021improved} & BanglaLekhaImageCaptions &0.651 &0.426 &0.278 &0.175 &0.572 &0.297 &0.434 & \textbf{\underline{0.357}} \\

\cite{humairahybridized} &Bengali Flickr4k-Bn & \textbf{0.653} & \textbf{0.505} &0.381 &0.226 &NA &NA &NA &NA\\
\\
\hline

\end{tabular}%
    \leavevmode
    {
     \small \\   $^{m}$Manual extraction of Arabic dataset \quad $^{x}$Subset of Arabic Flickr8k (2000 images) \\ Top performer in each language is in bold \quad  Top performer for each metric is underlined
    }
    

\end{table*}

\subsection{Hand-crafted approaches for image captioning}

Image captioning was initially performed using traditional machine learning techniques before advancement in deep learning methods \cite{kulkarni2013babytalk}. Pattern recognition systems have played an important role in solving computer vision tasks related to images \cite{kpalma2007overview}. Unsupervised and semantic segmentation approaches, which typically require less time and data than recent deep learning techniques, were commonly used \cite{sezgin2004survey}. In a study by \cite{jindal2017deep}, a three-stage root word-based method was proposed to generate Arabic captions for images. This involved creating image fragments using a pre-trained deep neural network on ImageNet and mapping them to a set of root words in Arabic. Furthermore, a deep belief network pre-trained by restricted Boltzmann machines was utilized to extract the most suitable words for the image \cite{almanaseer2021deep}.

\subsection{Deep learning for image captioning}

This section provides an overview of research papers that develop deep-learning approaches based on image captions.

\subsubsection{The root words recurrent neural network and deep belief network model}

In their work, \cite{jindal2018generating} proposed a method for generating captions directly from images in Arabic. They utilized root-word-based recurrent neural networks and deep neural networks. The process involved extracting root words from the images, translating them into morphological inflections, and then using the dependency tree relations of these words to establish the sentence order in Arabic. They used two datasets for their study: the Flickr8k dataset, which had manually written captions in Arabic by professional Arabic translators, and a collection of 405,000 images with captions from various newspapers in Middle Eastern countries. The findings indicated that the direct one-stage generation of Arabic captions yielded better results than a two-stage process involving using English captions in the Arabic translation.

\subsubsection{The convolutional neural network-gated recurrent units
encoder-decoder model}

To address the issues of exploding and vanishing gradients in RNN, a proposed method was introduced by \cite{do2020reference}. The model was built upon an encoder-decoder architecture, utilizing CNN for image description and GRU (gated recurrent units) for text generation. The GRU decoder utilized an image feature vector extracted by CNN and information from the scores of phrase weights. Two methods were applied to generate the scores. The first method used the part-of-speech (PoS) technique to produce scores based on word classes, while the second method utilized a likelihood function measured by the Euclidean distance. The results indicated that the PoS approach outperformed the model.

Furthermore, \cite{pedersoli2017areas} introduced an innovative approach that modeled the direct dependencies between caption words and image regions. This transformer-based approach could dynamically focus on various parts of the image. The proposed model included a CNN encoder to extract features from the image, and an RNN-based gated recurrent unit (GRU) was used as a decoder to simplify the model. The model was further enhanced by incorporating an attention mechanism to generate captions word by word for different image regions. This allowed the words to represent specific image regions rather than global areas, improving performance.

\subsubsection{The convolutional neural network-recurrent neural networks - long short-term memory encoder-decoder model}

The work of \cite{chen2019news} proposed a multi-modal attention mechanism for generating news image captions. This mechanism combines visual and textual attention to generate captions from news images and text. The goal is to ensure that the caption of a news image reflects the specific event reported, making it different from a general caption. More than 98\% attention was paid to the text, while the rest focused on the image.

The work of \cite{zakraoui2019improving} introduced a text-to-picture system comprising several steps: keyword extraction, query formulation, image selection, image captioning, sentence similarity, image ranking, and image evaluation. This work identified challenges in mapping natural text to multimedia, including a lack of captions and meaningful tags for images returned from the Google search engine. To address this issue, they proposed using a deep-learning captioning model.

In the attention mechanism, determining the optimal number of regions to capture all the details in an image can be challenging. To address this issue, \cite{biswas2020towards} proposed an approach that combines low- and high-level images. They used a combination of a Convolutional Neural Network (CNN) and an LSTM-based decoder to generate image captions. The visual attention mechanism is based on the history of image feature generation, and re-ranking methods were employed to measure the similarity between the generated captions and the corresponding object classes. 

The work of \cite{al2018automatic} introduced a generative merge model for Arabic image captioning. This model involves the interaction of two subnetworks to generate captions. The language model is based on RNN-LSTM to encode linguistic sequences of different lengths. At the same time, the image encoder is a fully convolutional network based on the Visual Geometry Group (VGG) that extracts image features as a fixed-length vector. A decoder model takes the fixed vectors from the previous models as input and makes the final prediction. It was suggested that this merged model could achieve excellent results for Arabic image captioning with a larger corpus.

In addition, \cite{mualla2018development} developed the Arabic Description Model (ADM) to generate full image descriptions in Arabic, compared to an earlier model based on English. The image features were obtained from CNN, and a JSON file containing image descriptions in English was translated into Arabic and fed to an LSTM network along with the CNN feature vector. The authors reported that translating recognized English captions into Arabic resulted in poor sentence structure, indicating that it is not a viable approach.

\begin{figure*}[ht!]
\centering
\includegraphics[width=11cm]{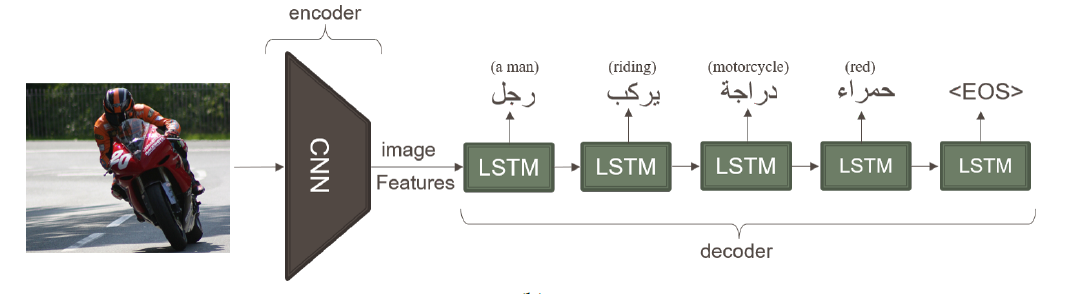}
\caption{Sequence-to-sequence encoder-decoder framework for Arabic language image captioning \cite{eljundi2020resources}. }
\label{Eljundi}
\end{figure*}

In addition, \cite{eljundi2020resources} developed a new Arabic image captioning dataset and evaluated two models with this dataset, demonstrating the superiority of the end-to-end model. Fig. \ref{Eljundi}  illustrates the proposed model employing a sequence-to-sequence encoder-decoder framework for image captioning. This involves encoding the input image into a feature vector using CNN and decoding that feature vector into an Arabic sentence using RNN.

An automatic model that converts standard Arabic children's stories into representative images that support the meaning of the words was proposed by \cite{saleh2019towards}. The method consists of seven steps: Keyword extraction, query formulation, image selection, captioning, sentence similarity, image ranking, and image evaluation. Teachers or parents can use this system to help children review the materials they have studied in school.

In a separate study, \cite{cheikh2020active} presented recent work on Arabic image captioning. Their research introduced an architecture-based encoder-decoder that outperforms classical methods using the standard Neural Machine Translation (NMT) approach. This approach used a CNN as an encoder to extract visual information from the input image. At the same time, an LSTM acts as a decoder, producing a probability distribution over possible next steps to generate the caption. The proposed active learning framework involved human annotators to refine the automatic translation produced by the model.

The automatic captioning of images in Indonesian was developed by \cite{mulyanto2019automatic}. The model comprises three components: an image extractor that generates feature vectors using CNN, a sequence processor that encodes linguistic sequences based on LSTM using the output from the previous step, and a decoder that predicts the caption of a new image based on the vector image features and vocabulary input. The test set showed promising results.

The work of \cite{tien2020image} introduced a new Vietnamese image captioning method. This method comprises an image captioning model, an English-Vietnamese translation model, and an unknown word processing model. The image feature extractor utilizes CNN, and the translation model comprises an encoder-decoder, RNN. Additionally, the model provides an unknown word processing module to address the problem of unknown words in Vietnamese translation.

Myanmar's \cite{pa2020automatic} proposed an image captioning method combining two parts: CNN for image feature extraction and LSTM for text generation. New datasets were built based on the Flickr8k dataset. The new datasets used 3,000 images from the Flickr8k dataset, each with five annotated Myanmar captions. This approach reduced the manual captioning time by translating the sentences. The generated text was evaluated using BLEU, and satisfactory results were obtained.

A new model for Bengali image captioning was proposed by \cite{humairahybridized}. The model utilized two-word embedding techniques and consisted of a two-part encoder and decoder. The encoder comprised a convolutional neural network, while the decoder included BiLSTM and BiGRU. The process involved extracting the image features and concatenating the output word vectors, which were then passed to the decoder after aligning the dimension between the word vector and the image features. The decoder utilized the concatenated output to generate the next word in the sequence with the highest probability. The Flickr8k dataset was used for testing, with five captions for each image translated into Bengali using Google Translator.

Bengali Image Captioning (BIC) was also presented in \cite{khan2021improved}. The model consisted of an image feature encoder, a word sequence encoder, and a caption generator. The model was tested on the BanglaLekhaImageCaptions dataset, which contains 9,154 images, each with two captions generated by two native Bengali speakers.

\subsubsection{A summary of deep learning methods}

In most methods, the image is first fed into a CNN to generate image features, which will then be used as input for the language processing component. The convolution layer reduces the image into features by using information from nearby pixels. It then employs prediction layers to forecast the target values. This is achieved by creating a dot product using multiple convolution filters, or kernels, which scan the image and extract unique aspects of the image. The max pooling layer helps to reduce the spatial size of the convolved features and prevents overfitting by providing an abstract representation of the convolved features. Although there are many different activation functions, RelU is the most commonly used one in various types of neural networks due to its ease of training and superior performance due to its linear behavior \cite{alzubaidi2021review}.

\begin{figure*}[ht!]
\centering
\includegraphics[width=10cm]{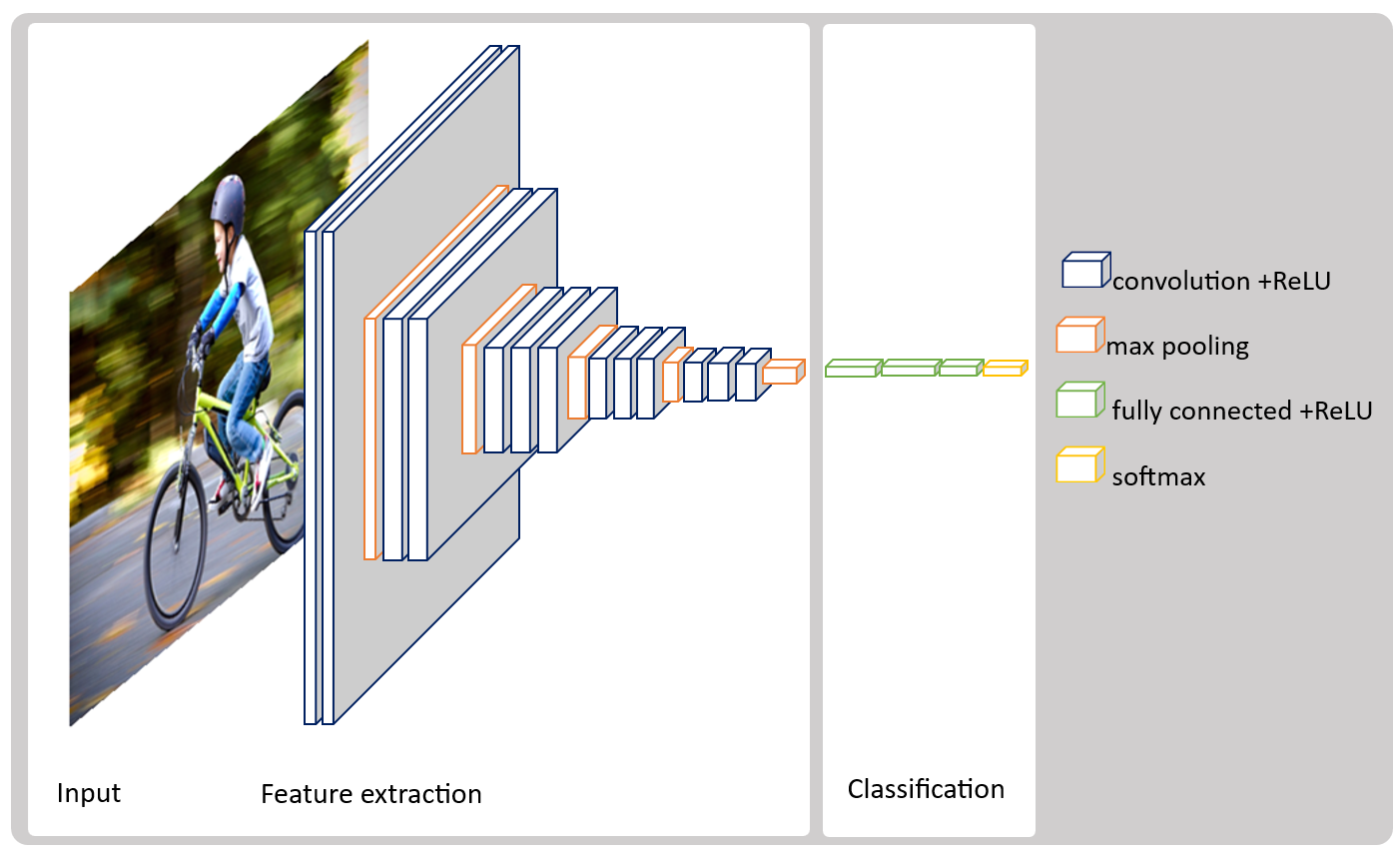}
\caption{A typical architecture of the CNN model.}
\label{cnn}
\end{figure*}

In a CNN network, the higher layers are believed to capture high-level semantic information. As a result, the output of the fully connected layer can represent the image's global information. However, since this output lacks spatial information, the output of the last convolutional layer is often utilized. This is because the expanded receptive field of the higher layer in CNN corresponds to a region of the original image, where each point on the spatial feature map corresponds to a region of the original image \cite{zhang2019image}. The architecture of the CNN model is shown in Fig. \ref{cnn}.

\begin{figure*}[ht!]
\centering
\includegraphics[width=10cm]{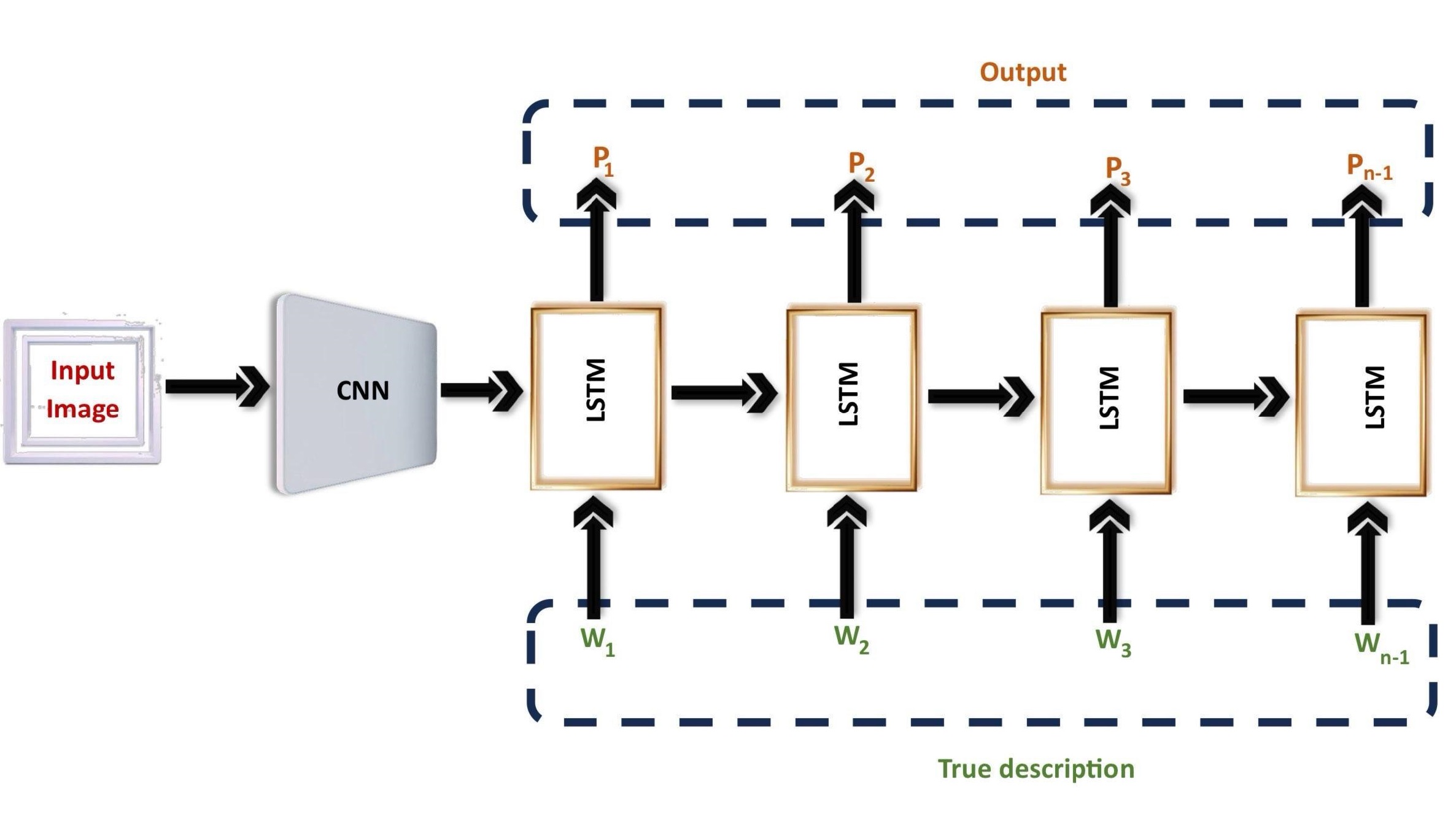}
\caption{A general architecture of the CNN-LSTM encoder-decoder model for image captioning.}
\label{CNNLSTM}
\end{figure*}

The general architecture of image captioning models that use the encoder-decoder framework is depicted in Fig. \ref{CNNLSTM}. The encoder comprises a CNN for extracting image representations, while the decoder incorporates an LSTM for generating image captions.
CNNs are a type of feedforward artificial neural network that is adept at processing visual data. A typical CNN consists of an input, an output, and multiple hidden layers. The hidden layers of a CNN typically include convolutional, pooling, fully connected, and normalization layers.
On the other hand, text generation is handled by an essential deep learning model capable of learning long-term dependencies, the LSTM. An LSTM consists of a cell, an input gate, an output gate, and a forget gate as its internal components. Using simple learned gating functions, the internal units of an LSTM utilize nonlinear mechanisms to enhance hidden states, allowing them to propagate unchanged, be updated, or be reset.

\subsection{Transformers-Based Approaches for Image Captioning}

This section reviews research papers that focus on developing transformers that generate captions.

\subsubsection{CNN-transformer encoder-decoder model}

The study by \cite{yu2019multimodal} introduced a multi-transformer (MT) for image captioning. This MT model can understand three types of relations: word-to-word, object-to-object, and word-to-object. The transformer mechanism consists of an image encoder and a text decoder. The image encoder has two parts: an aligned multiview encoder and an aligned multi-view decoder. The caption decoder takes the output from the encoder and generates a caption using word embedding and one layer of LSTM.

In a different approach, \cite{wang2020transformer} utilized a faster region-CNN (R-CNN) to extract visual features for a given image. These features are then inputted into the transformer encoder, allowing the transformer to effectively capture object information by overcoming interference from non-critical objects. The attention matrix computed from the transformer encoder is passed into the attention gate, where the attention weight values below the gate threshold are truncated. The decreasing threshold leads to the preservation of more non-zero values, expanding the attention scope of the self-attention module from local items to all objects as the network layer expands.

In traditional practices, normalization has been applied outside of self-attention. However, a study by \cite{guo2020normalized} introduced a novel normalization method and demonstrated its feasibility and advantages for hidden activation within self-attention. They proposed a geometry-aware self-attention (GSA) class that extends self-attention to explicitly consider relative geometry relations between objects in an image for feature extraction.

The work in \cite{kumar2022dual} employed a dual-modal transformer to capture intra- and inter-model interactions within an attention block. They concatenated two embeddings, one based on the image's objects and the other using an Inception-V3 model, to create the final image-based embedding. The study showed that this model outperformed established models such as encoder-decoder and attention models.

State-of-the-art techniques directly encode identified object regions and utilize region features. However, this approach presents challenges related to object relationships and the potential for incorrect item detection. Significant computational power is also required to compute region features, particularly when using high-performing CNN-based detectors like Faster R-CNN. The study by \cite{nguyen2022grit} addressed these issues by replacing a CNN backbone with a transformer to overcome the drawbacks of CNN-based detectors and reduce computational costs for extracting initial features from input images.

\subsubsection{LSTM-transformer encoder-decoder model}

In their work, \cite{he2020image} introduced a new transformer-based model that considers the relationships between different features within an image. This model considers three types of spatial relationships in the image regions: a query region can be a parent, neighbor, or child. The model uses spatially adjacent matrices to combine the output of parallel subtransformer layers. The decoder includes an LSTM layer and an implicit transformer layer, which work in parallel to decode different image regions.

Two new geometry-aware architectures were separately created for the encoder and decoder to represent geometry better \cite{wang2022geometry}. This captioning model helps us understand the locations of target objects and the objects the model is currently looking at. The proposed model includes an improved encoder and may provide information on an object's relative geometry. Furthermore, it fully leverages geometry relations to enhance object representations.

Remote sensing image captioning (RSI) aims to generate descriptions of the information contained in RSIs automatically. The multiscale information of RSIs encompasses the properties and complex relationships of items of various sizes. The study by \cite{liu2022remote} developed a new model based on the encoder-decoder framework. In this model, ResNet50 serves as the encoder to extract multi-scale information. At the same time, a multi-layer aggregated transformer (MLAT) is employed in the decoder to construct sentences using the extracted data effectively. Additionally, LSTM aggregates features from multiple transformer encoding levels to enhance feature representations.

\subsubsection{Transformer-based model}

In their work, \cite{li2019entangled} proposed a novel transformer-based approach to address the limitations of recurrent neural networks (RNN). They introduced an attention mechanism that combines visual and semantic attention to handle complex relationships. Since not every word has a corresponding visual signal, taking into account semantic information is crucial. The proposed method includes a control mechanism for the forward propagation of multi-model information. The model utilizes a dual-way transformer encoder to investigate inter- and intra-relationships between visual and semantic attributes. The decoder's output is passed to a classifier to predict the next word.

In \cite{fei2022attention}, a transformer-based model was introduced for image captioning. The approach involved using a mask operation to automatically evaluate the impact of the features of the image region and using the results as supervised information to guide attention alignment. The basic version of the transformer was utilized in this study. The researchers investigated the relationship between attention weights and feature importance metrics in image captioning to comprehensively analyze whether current attention mechanisms can focus on crucial and effective image regions. This work serves as a valuable reference for self-supervised learning.

The research by \cite{WEI2022104574} introduced an attention-based approach. The model is designed to capture dependencies within image areas and between image regions and external states. Using the self-attention method, the captioning model can identify the most relevant regions at each time step. The researchers explored a sequential transformer framework based on the original transformer structure, combining the decoder with outside-in attention and RNN. The study revealed that the transformer's self-attention allows for the simultaneous direct calculation of relationships between internal areas, thus avoiding recurring attention issues.

The LATGeO framework, based on transformer technology, was introduced in \cite{dubey2023label} to generate captions for images. It incorporates multi-level geometrically coherent and visual recommendations to establish relationships between objects based on their localized ratios. A new label-attention module (LAM) was developed to connect the visual and linguistic aspects to extend the traditional transformer. In this proposed approach, object labels are included as input data at each decoder layer to assist in constructing captions.

In the context of image captioning, \cite{jiang2021multi} proposed a Multi-Gate Attention (MGA) block within a pre-layer norm transformer architecture. This architecture modifies the standard self-attention mechanism by incorporating multiple gate mechanisms, thus enhancing its capabilities. The pre-layer norm transformer architecture differs from the original transformer architecture in that the layer normalization is placed before the self-attention module, and the feedforward layer and subsequent layers are eliminated. This simplification aims to increase the model's efficiency for image captioning.

The Transformer architecture, which was recently announced, utilizes self-attention to enhance the performance of sequence-analysis tasks. This has led to exploring transformers in \cite{kandala2022exploring}. The experimental validation was conducted using the caption dataset from the University of California (UC)-Merced. The proposed technique can potentially generate helpful textual descriptions for remote-sensing images.

In transformer-based image captioning, three-parameter reduction techniques were utilized \cite{tan2022acort}. Firstly, the size of the embedding matrices was significantly reduced by using radix encoding, allowing for a larger vocabulary without increasing the model size. Secondly, cross-layer parameter sharing was employed to break the tight correlation between model depth and size, allowing additional layers to be added without increasing the parameter count and vice versa. Finally, attention parameter sharing was used to reduce the parameter count of the multi-head attention module and improve overall parameter efficiency.

To effectively capture complex interactions within and between input features in images, a Modular Co-Attention Transformer Layer (M-CATL) was proposed by \cite{wang2022dm}. This layer aims to extract specific image characteristics. Furthermore, a Deep Modular Co-Attention Transformer Block (DM-CATB) was developed and integrated into the encoder part of the model based on M-CATL. To fully capture spatial and positional information of image features and improve feature characterization, a Deep Modular Co-Attention Transformer Network (DM-CATN) was introduced.

Local visual modeling with grid features is crucial to generating comprehensive and detailed image captions. In their work, \cite{ma2023towards} proposed a locality-sensitive transformer network (LSTNet) to facilitate local object recognition during captioning. They also employed layer-specific fusion (LSF) for cross-layer semantic complement, combining information from multiple encoder layers. The experimental results demonstrated that LSTNet's local visual modeling outperformed many state-of-the-art captioning models.


The study by \cite{cornia2020meshed} introduced an improved architecture for image captioning by incorporating a unique memory mechanism into a Transformer-based framework, addressing the challenges of maintaining long-range relationships and contextual coherence in traditional image captioning algorithms. The authors proposed the Meshed-Memory Transformer (MMT), which integrates a memory module to improve the model's capacity to retain and utilize data in both temporal and spatial dimensions. This memory-enhancing mechanism and a typical Transformer model helped the MMT system capture complex links between generated text and visual elements, leading to more detailed and cohesive captions. The research demonstrated that MMT significantly improved captioning performance in various benchmark datasets.

Additionally, a technique called the full memory transform was described in the work by \cite{lu2023full}. This technique aims to enhance the efficiency of language decoding and image encoding. The Full-Layer Normalization Symmetric Structure for Image Encoding was suggested, embedding Layer Normalization symmetrically on both sides of the self-attention network (SAN) and feedforward network for robust training and higher model generalization performance. Furthermore, the Memory Attention Network was introduced to extend the conventional attention mechanism, directing the model to concentrate on words that require attention, thus improving the language decoding step.

\subsubsection{Summary of transformer-based models}

\begin{figure*}[ht!]
\centering
\includegraphics[width=13cm]{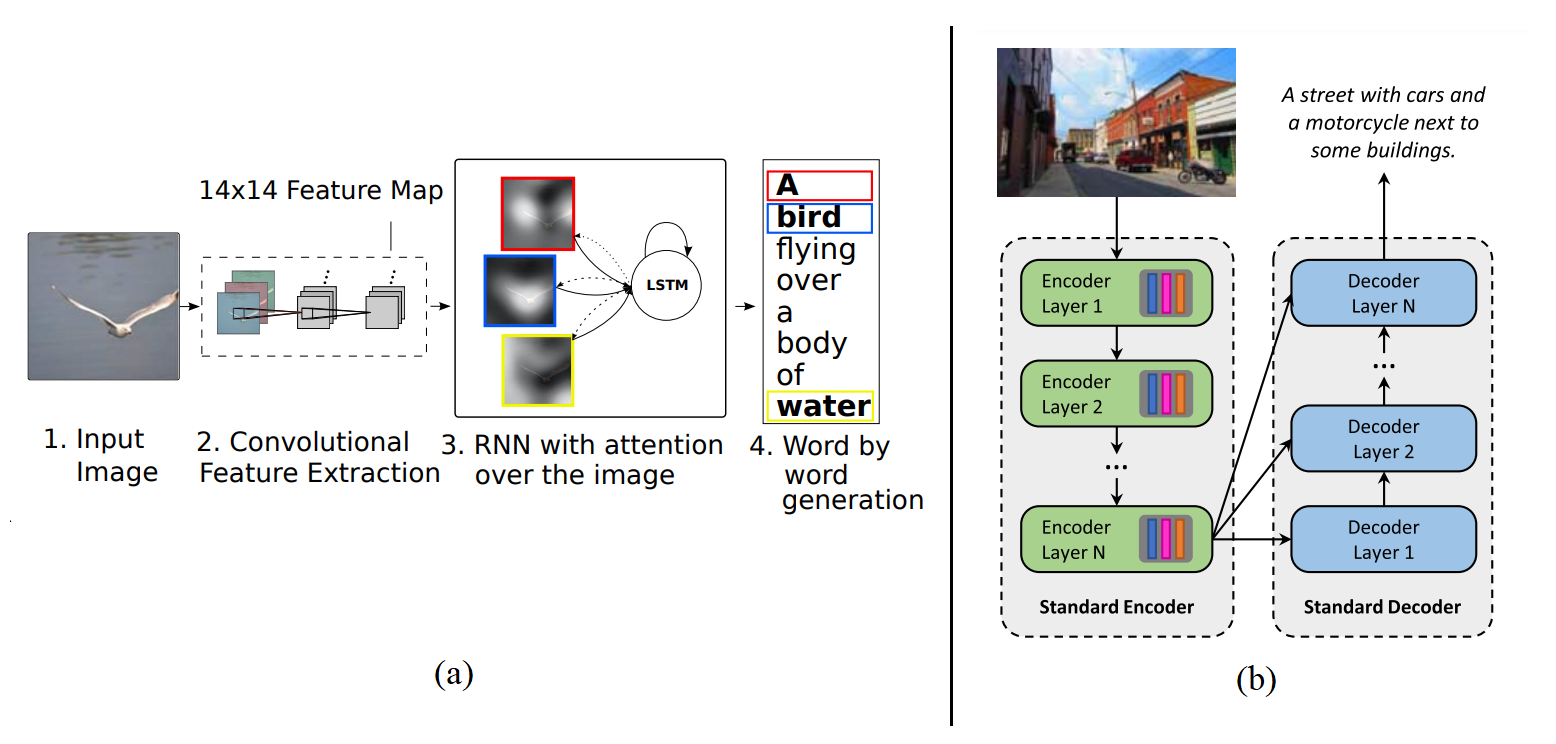}
\caption{Image captioning models: (a) CNN-RNN based model \cite{xu2015show}, (b) transformer-based model \cite{cornia2022explaining}.}
\label{rnntransformer}
\end{figure*}

Transformers generate the words of the caption all at once, while model-based RNN still produces the caption word by word (see Fig. \ref{rnntransformer}). In the model on the left, which is a CNN-RNN-based model, the caption words are produced one by one \cite{xu2015show}. On the other hand, the model on the right demonstrates the transformer's ability to generate a full text with all words simultaneously \cite{cornia2022explaining}.

\begin{figure*}[ht!]
\centering
\includegraphics[width=10cm]{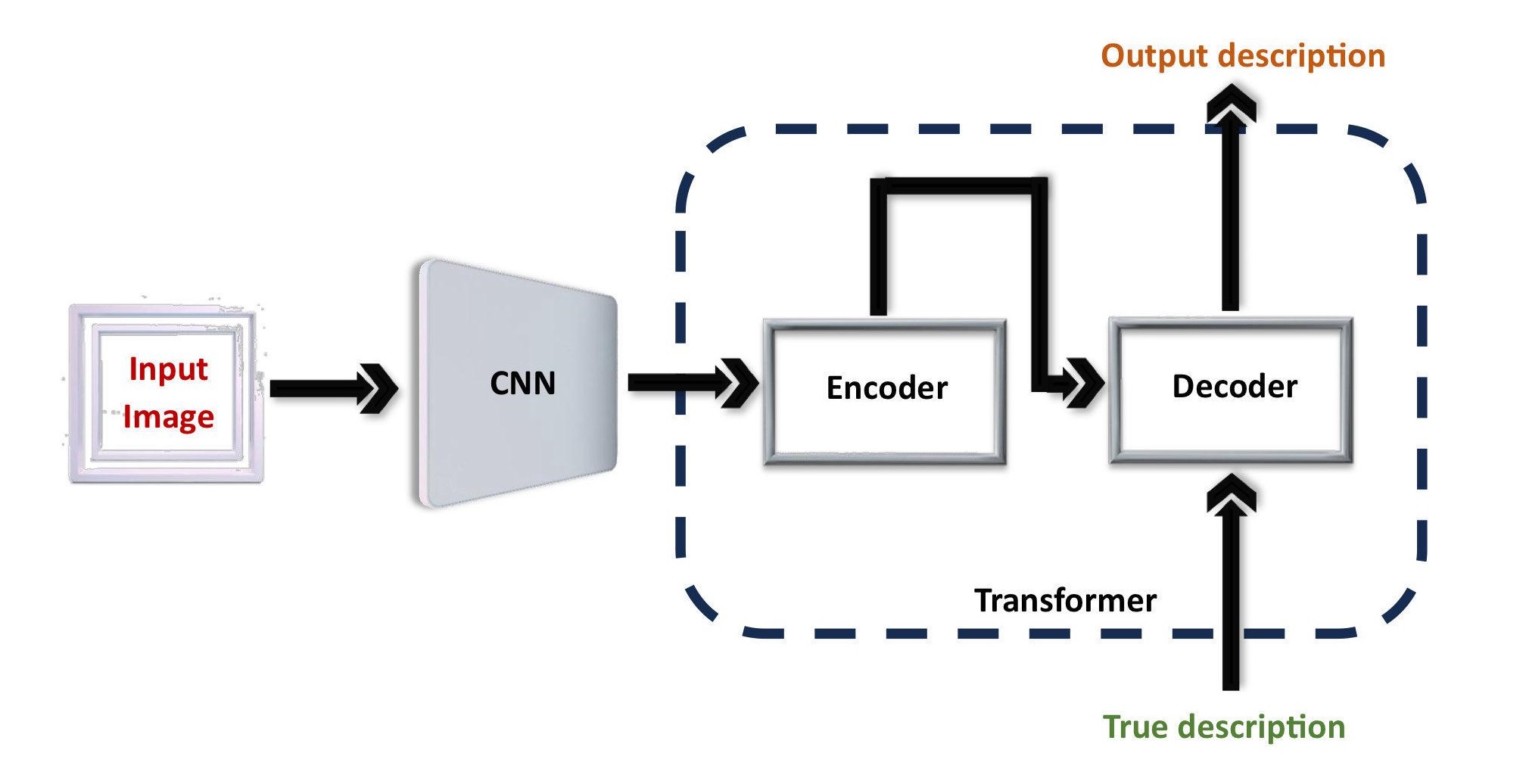}
\caption{General architecture of CNN-transformer model for image captioning.}
\label{CNNTrans}
\end{figure*}

Fig. \ref{CNNTrans} shows a general architecture of the transformer model for image captioning. It includes a feature extraction model, typically a CNN, and a transformer for text generation. The transformer comprises an image encoder to learn self-attended visual features and a caption decoder to generate the caption from the attended visual features.

\subsection{Attention-based approaches for image captioning}

The study by \cite{pan2020x} introduced a new approach to address the computational limitations of traditional attention mechanisms in image captioning. The authors presented X-Linear Attention Networks (X-LAN), which combine a linear attention module for improved computational efficiency and reduced complexity with a non-linear module capturing more detailed interactions and dependencies within the image.
The study demonstrated that X-LAN produces significant performance improvements in benchmark datasets compared to existing methods, offering a more scalable and effective solution to generate detailed and contextually accurate image descriptions. By enhancing both efficiency and accuracy, X-LAN advanced the capabilities of image captioning systems.

A separate study by \cite{huang2019attention} improved the attention mechanisms used in image captioning. They proposed a novel framework called Attention on Attention (AoA), which enhances existing models by introducing a secondary attention mechanism that acts on the primary attention outputs. This secondary process reassesses and recalibrates the original attention weights, considering the generated words' context and visual elements' context.

In their work, \cite{ke2019reflective} proposed a new method called Reflective Decoding Network (RDN) to enhance image captioning systems. Unlike traditional models, which often employ a single-stage decoding process that may not fully utilize the context and finer details of the visual input, RDN involves a two-step decoding process. In the first stage, a reflective mechanism generates an initial caption, followed by a second stage to refine it. This reflective decoding process employs a self-attention-based approach to review and modify the original caption, considering the visual elements and previously generated words. This iterative refinement results in improved captioning output from the model.

Developing non-visual words such as "to" and "itself" does not require much visual information. Therefore, using image features as key-value pairs for cross-attention to create captions for images is unsuitable. In the Task-Adaptive Attention model proposed in \cite{9381876}, task-adaptive vectors were included to learn nonvisual signals that can help address this issue in image captioning. The comprehensive Transformer model with Task-Adaptive Attention integrates the suggested task-adaptive attention module into a standard transformer-based encoder-decoder architecture.

In a study by \cite{wang2021reasoning}, a novel image captioning technique called Dynamic Attention Prior (DY-APR) was introduced. This approach combines attention distribution before the local linguistic context for dynamic attention aggregation. The researchers proposed a method for dynamically aggregating the Attention Distribution Prior (ADP) and the current layer's attention score to provide more precise attention guidance. They also presented a learning technique to gradually transition input tokens from a fully static representation based on word embedding to a mixed scheme incorporating both the input tokens and the linguistic context.

Existing image captioning methods focus primarily on the visual attention mechanism, often resulting in incomplete and inaccurate model-generated sentences. In addition, errors in extracting visual features can lead to incorrectly generated captions.  The work of \cite{liu2022image} addressed this gap by proposing a combination attention module consisting of two modules: visual attention and keyword attention. The evaluations demonstrated that this strategy yielded better results.

\subsubsection{Soft and hard attention}

The first attentive deep paradigm for image captioning was Show, Attend, and Tell \cite{li2019show}. In this model, the decoder used an LSTM for language modeling, and the feature extractor was a CNN. Specifically, the VGG model was pre-trained on ImageNet. Show, Attend, and Tell was quite similar to other CNN-LSTM encoder-decoder architectures for captioning videos, except that it utilized two attention mechanism variants: soft and hard attention on the spatial convolutional features to generate a set of attended features for the LSTM decoder, acting as a language model.

In Show, Attend, and Tell, attention involves a set of attended visual features ($z$) generated from an attention function (fatt), which can be soft deterministic or hard and stochastic. In soft attention, the input to the LSTM comprises weighted image characteristics that take attention into account rather than the image $x$. Soft attention reduces the weight of irrelevant places with low attention, which helps to focus on relevant areas.

In soft attention, areas of high focus retain their original values, while areas of low focus approach 0. This is achieved by assigning a weight, $a_i$, to each $x_i$ input to the LSTM. The sum of all weights, $a_i$, is 1, representing the likelihood of focusing on $x_i$. On the other hand, hard attention uses a stochastic sampling model by selecting $x_i$ as input to the LSTM, with $a_i$ serving as a sampling probability rather than a weighted average.

The Monte Carlo approach is used in hard attention to accurately calculate the gradient descent during backpropagation, while soft attention uses the standard backpropagation method \cite{li2019show}. This allows the model to concentrate its computation on specific salient regions while generating captions, using soft and hard attention to understand the concept of attention in image annotation.

Soft attention can be trained through standard backpropagation by applying weights to the annotated vector of picture features when the feature is salient. In contrast, stochastic hard attention can be trained by maximizing the lower bound variation \cite{oluwasammi2021features}.

\begin{figure*}[ht!]
\centering
\includegraphics[width=8cm]{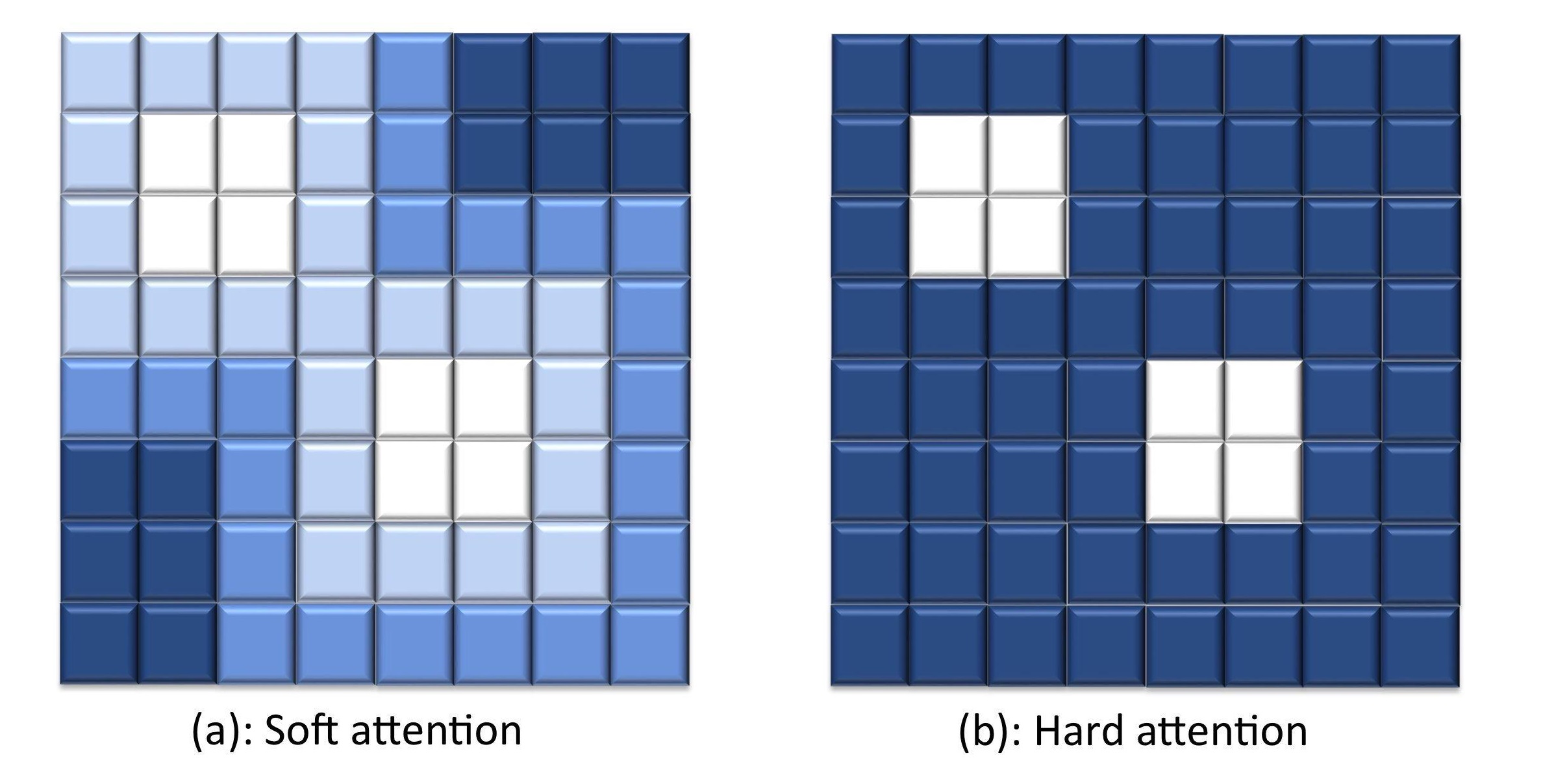}
\caption{Examples of soft and hard attention mechanisms. (a) Soft attention assigns varying importance to different parts of the image, influencing the entire caption. In contrast, (b) hard attention focuses on specific regions, selectively considering parts of the image while ignoring others.}
\label{soft_and_hard}
\end{figure*}

It is important to note that the spatial information extracted from the two-dimensional image is crucial for both soft- and hard-attention mechanisms. The extracted annotation vector contains features of each color channel in a 3-dimensional spatial feature vector since the images are represented using three color channels (red, blue, and green). Once the set of spatial features attended is determined, they are ready for use \cite{zohourianshahzadi2022neural}. Soft and hard attention can be seen in Fig.\ref{soft_and_hard}. In this illustrative figure, you can observe how soft attention demonstrates the relative importance of each part of the image to the other parts. In contrast, hard attention separates specific parts of the image and considers only these parts when generating the next word in the caption, disregarding the rest \cite{xu2015show}.

\subsubsection{ Bottom-up and top-down approaches}

Employing saliency is based on how our brain processes visual information. It combines a bottom-up flow of visual inputs with a top-down reasoning process. The top-down approach involves predicting incoming information from the environment using our prior knowledge and logical bias. In contrast, the bottom-up approach involves visual signals that correct the prior predictions. Additive attention can be approached as a top-down system, where the language model observes a feature grid independent of the image content and predicts the subsequent word. The bottom-up path is defined by an object detector responsible for identifying image regions. Then, a top-down process learns to weigh each region for each word prediction. This approach is connected with the concept of additive attention \cite{stefanini2022show}.

Scientists have focused on top-down and bottom-up attention theories, and recent research has shown that top-down attention mechanisms are still preferable. The top-down model starts with an image as input and converts it to words \cite{staniute2019systematic}. All methods that utilize bottom-up attention perform better because bottom-up attention focuses on visual attention at the object level. However, an important question arises: In some natural settings, is it necessary for the model to pay attention to areas in the image that do not contain recognizable objects but instead include natural elements such as mountains, trees, skies, etc.? On the other hand, using object detectors for bottom-up feature extraction has drawbacks, as they may not be able to focus on important areas for captions in unfamiliar domains. As a result, additional knowledge from various domains may be necessary for natural settings, and object detectors trained in more specialized tasks can provide this kind of knowledge \cite{zohourianshahzadi2022neural}.

Attention-based encoder-decoder models are known for their sequential information processing but are criticized for lacking global modeling skills. To overcome this limitation, a reviewer module has been developed to conduct review stages on the encoder's hidden states and generate a thought vector at each step. The attention mechanism achieves this by assigning weights to the hidden states. The thought vectors capture global aspects of the input and effectively review and learn the information encoded by the encoder. The decoder uses these thought vectors to predict the next word \cite{bai2018survey}. Additionally, incorporating visual attention allows for a multimodel coverage mechanism \cite{chen2019news}. This visual attention mechanism uses features derived from a convolutional neural network layer, where each feature represents an abstraction of a region in the image and provides a weighting for each geographical region. A higher weight indicates a more important image region \cite{biswas2020towards}. It is worth mentioning that the described attention method falls between the encoder and the decoder.

\begin{figure*}[ht!]
\centering
\includegraphics[width=10cm]{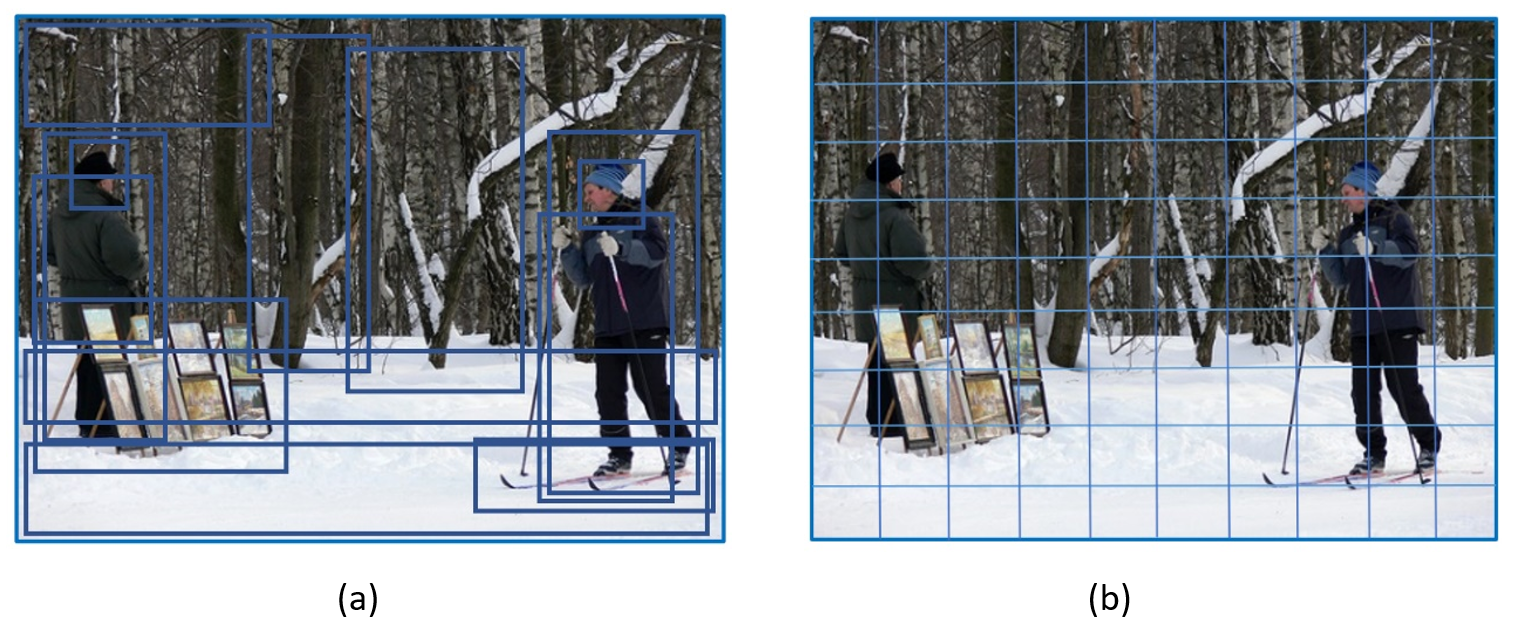}
\caption{Illustrations of bottom-up and top-down attention approaches. (a) Bottom-up attention, where focus is determined at the level of objects and other salient regions of the image, and (b) top-down attention, where features correspond to a uniform grid of equally sized image regions.}
\label{bottom-up_and_top-down}
\end{figure*}

Figure \ref{bottom-up_and_top-down} illustrates an example of bottom-up and top-down approaches. A set of salient image regions is identified in the bottom-up approach, and a pooled convolutional feature vector, like Faster R-CNN, an exemplary bottom-up attention mechanism, describes each region. The top-down approach, on the other hand, utilizes the task-specific context to determine an attention distribution over the visual regions. The weighted average of the image features in all regions is then utilized to compute the attended feature vector.  The study by \cite{anderson2018bottom} proposed a method that presented bottom-up and top-down approaches.

\subsubsection{Summary of attention-based models}

\begin{figure*}[ht!]
\centering
\includegraphics[width=9cm]{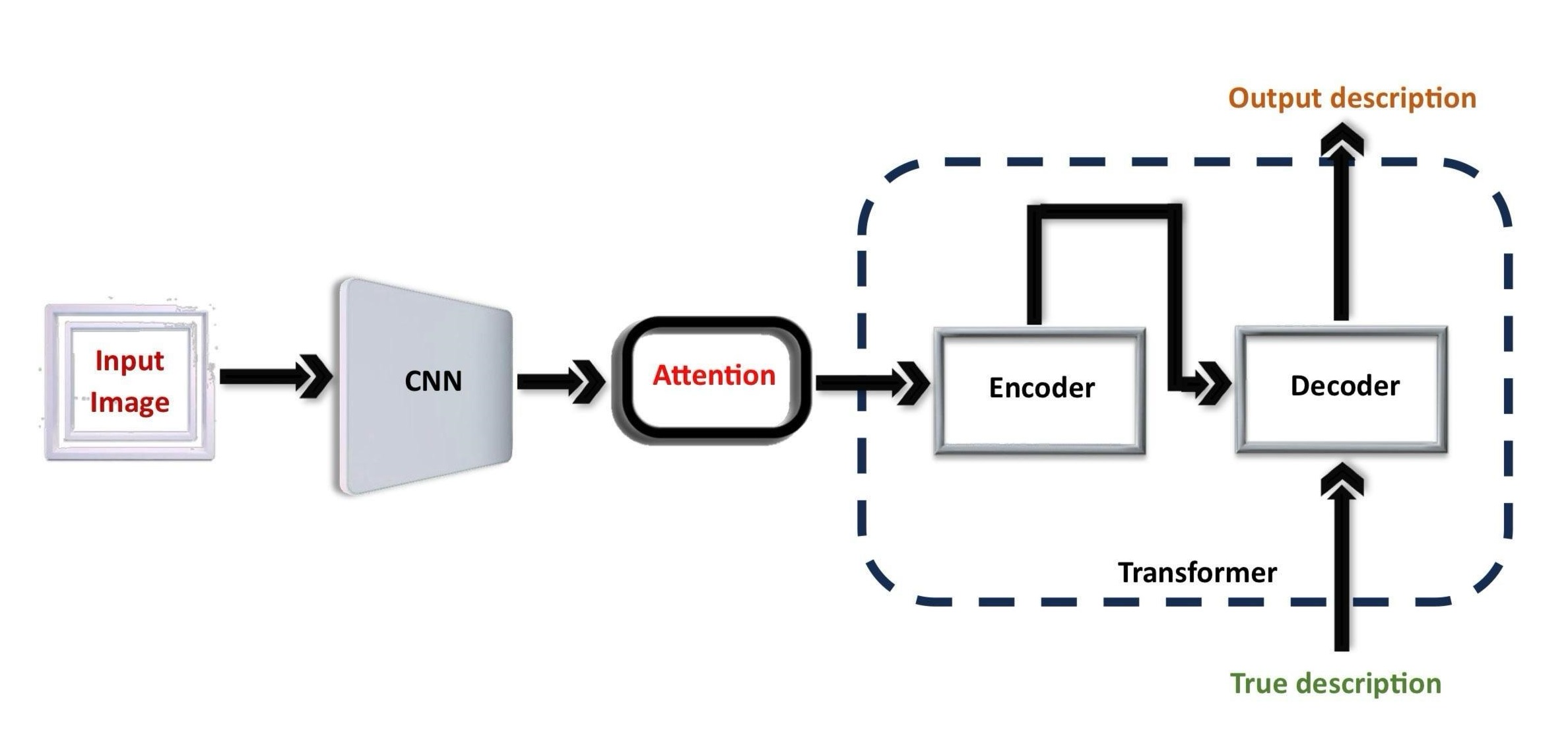}
\caption{General architecture of an attention-based model for image captioning, illustrating the integration of image features with sequential attention mechanisms to generate descriptive captions.}
\label{CnnAttTrans}
\end{figure*}

Figure \ref{CnnAttTrans} illustrates the general architecture of the attention model. This innovation has greatly enhanced image captioning, allowing the algorithm to focus on important image aspects and ignore redundant content. This model implements attention as a weighted sum of encoder outputs. A CNN first processes the image within the encoder-decoder framework, resulting in feature maps. Subsequently, the attention module assigns a weight to each image pixel based on the feature maps and a hidden state. These weights enable the decoder to generate words for the output text while concentrating on the most pertinent parts of the image.
 
\subsection{Graph-based representation for image captioning}

The study by \cite{yao2018exploring} emphasized the importance of visual relationships among objects, advancing the field of image captioning. Traditional image captioning models typically focus on object detection and identification, generating descriptive text based solely on these aspects. However, such approaches often neglect the intricate connections and interactions between objects that can greatly enhance the depth of the captions. The authors introduced a new approach integrating a visual relationship module into the captioning architecture to address this limitation. This module analyzes and encodes the interactions between elements in an image using a graph-based representation. This enables the model to understand better and express the spatial and functional relationships between items, resulting in more detailed and contextually accurate captions. The research offered a comprehensive analysis of their methodology, demonstrating significant improvements in relevance and caption quality compared to existing approaches. The authors expanded the boundaries of current image captioning systems by showcasing through extensive experiments that incorporating visual relationships enhanced the descriptive power of the captions and improved the model's ability to generate coherent and contextually appropriate descriptions.

\subsection{Comparative analysis of state-of-the-art methods for image captioning}

This section evaluates the effectiveness of various state-of-the-art methods for Image Captioning, as presented in Table \ref{tab:results}. The table includes numerous methods used on different datasets, including Arabic datasets such as Flickr8k and Flickr30k and English datasets like Flickr8k, Flickr30k, and MSCOCO, along with other datasets from various languages. Performance was measured using several metrics, including CIDEr, METEOR, ROUGE-L, SPICE, and BLEU scores at four levels (BLEU-1 through BLEU-4). Upon a thorough examination, it is evident that the technique \cite{jindal2018generating} achieved a high BLEU-1 score of 0.658 when applied to the Flickr8k Arabic dataset using manual extraction. Compared to non-English datasets, methods applied to English-based datasets like Flickr8k and Flickr30k generally yield better scores across most criteria.

In the English Flickr8k dataset, the method described in \cite{jiang2019modeling} achieved a high BLEU-1 score of 0.690, giving a strong performance in generating relevant captions. However, with the Arabic Flickr8k dataset, the method in \cite{elbedwehy2023improved} only achieved a BLEU-1 score of 0.598, revealing the challenges of adapting methods to different languages and contexts. Additionally, when the method described in \cite{humairahybridized} was applied to the Bengali Flickr8k dataset, it produced lower scores, with many values marked as 'NA,' suggesting that all metrics did not evaluate the approach.

Further analysis shows that methods evaluated using the English MS COCO dataset, including the method \cite{Wang2022EndtoEndTB}, generally achieve high scores on various metrics, such as a CIDEr score of 1.360 and a BLEU-4 score of 0.414. This indicates that the MS COCO dataset, a widely used and comprehensive dataset, provides a reliable standard for evaluating image captioning methods.

Furthermore, with the MS COCO dataset, techniques such as \cite{yang2022reformer} and \cite{fei2022attention} demonstrate strong performance, scoring highly in the BLEU and CIDEr metrics, showcasing their effectiveness in generating diverse and accurate captions. However, achieving high performance on non-English datasets such as Bengali BORNON and Indonesian FEEH-ID is more challenging, underscoring the need for further research and development in multilingual and culturally diverse image captioning techniques. This variation underscores the importance of creating more inclusive datasets and methods that perform well in linguistic and cultural contexts.

The variation in results between metrics for the same approach suggests that no single measure can fully evaluate the effectiveness of image captioning methods. For example, a method might have a high BLEU score but a low CIDEr or SPICE score, indicating different strengths and weaknesses. This highlights the need for robust and adaptable methods for diverse datasets and languages. Comparative analysis underscores the importance of adapting image captioning methods to specific contexts. Although some methods perform well on certain datasets, they may not perform as well on others, emphasizing the importance of continued development and adaptation in image captioning.

\section{ Datasets}

This section introduces the commonly used datasets in image captioning. Table \ref{tab:dataset} illustrates the datasets' details.

\subsection{English datasets}
\subsubsection{The Flickr8K dataset}
The Flickr8K dataset, developed by \cite{hodosh2013framing}, was publicly released in 2013. It comprises 8,000 images sourced from the Flickr image-sharing platform. Compared to MS COCO, Flickr8K is relatively tiny and primarily contains photographs of humans and animals. The image descriptions were manually annotated using Amazon Mechanical Turk, with each image paired with five descriptive sentences, ensuring linguistic diversity in the captions.

\subsubsection{The Flickr30k dataset}
The Flickr30K dataset \cite{young2014image} is an expanded version of Flick-r8K, containing 31,783 captioned images. Each image is accompanied by five descriptive sentences, providing a diverse linguistic representation. The dataset primarily consists of photographs depicting people engaged in everyday activities and events, making it a valuable resource for training and evaluating image captioning models.

\subsubsection{The Microsoft COCO datasets }
The Microsoft COCO (MS COCO) dataset \cite{lin2014microsoft} is a large-scale benchmark widely used in image recognition, object detection, semantic segmentation, and image captioning. Each image is manually annotated via Amazon Mechanical Turk and includes objects from over 100 categories, representing real-world scenes with natural backgrounds. The dataset contains 82,783 training images, 40,504 validation images, and 40,775 test images with undisclosed labels. Each image is paired with five descriptive captions, making MS COCO a key resource for evaluating image captioning models.

\subsection {Arabic datasets}
The model proposed by \cite{al2018automatic} was trained and tested using images from the MS COCO and Flickr8K datasets. The MS COCO dataset contains over 330,000 images and 2.5 million captions, covering 80 object categories. The CrowdFlower crowdsourcing service was used to generate Arabic captions, resulting in 5,358 captions for 1,166 images from the training set, with an average of 4.6 captions per image.

The Flickr8K dataset, which consists of 8,000 images, initially includes five English captions per image. The first 2,261 images from its training set were selected for this study, and a professional translator created 750 Arabic captions. The remaining images were translated into Arabic using Google Translate, followed by manual verification by native speakers. In total, 3,427 images (from both MS COCO and Flickr8K) were used, with a vocabulary size of 9,854 words, and the longest caption containing 27 words. For experiments, the dataset was split into 2,400 training images (70\%), 411 development images (12\%), and 616 test images (18\%).

Two test scenarios were proposed in \cite{mualla2018development}: (1) \textit{Machine translation approach} -- English image descriptions were translated into Arabic using Google Translate, often leading to grammatical errors and poorly structured sentences. (2) \textit{LSTM-based Arabic generation} -- Instead of relying on translation, an LSTM-based Arabic language model was trained to generate more natural and grammatically accurate captions.

To evaluate the model’s performance, three distinct test sets were used:
(1) \textit{Simulated trained images} -- The model was tested using trained images from Flickr8K. (2) \textit{Unseen test images} -- The model was evaluated using images from the Flickr8K test set that were not seen during training. (3) \textit{Tishreen University dataset} -- A new test dataset was created using images from Tishreen University, further improving the reliability of the experiments.


The work by \cite{eljundi2020resources} introduced the Arabic Flickr8K dataset. To create this dataset, the original English Flickr8K dataset was translated into Arabic using a two-phase process: First, the Google Translate API produced an initial Arabic translation. Next, qualified Arabic translators carefully reviewed and refined the translations. After verification, the top three translations per image were selected from an initial pool of five. The final Arabic Flickr8K dataset consisted of 6,000 training images, 1,000 validation images, and 1,000 test images, each with three unique captions.

In a separate study, \cite{cheikh2020active} developed the ArabicFlickr1K dataset using an active learning-based framework to translate an existing dataset. The final ArabicFlickr1K dataset contains 1,095 images, with three to five Arabic captions per image. This dataset was designed to support Arabic image captioning models, offering a diverse and linguistically rich dataset for improved training and evaluation.

\subsection {Other languages datasets}
The Vietnamese image captioning model proposed by \cite{tien2020image} was evaluated using the UIT-ViIC dataset, which was carefully curated to ensure consistent and accurate captions. The annotation process was conducted by five native Vietnamese speakers (aged 22–25) trained in sports-related vocabulary before starting. The dataset consists of 3,850 sports-related images sourced from the 2017 Microsoft COCO edition, with each image accompanied by five Vietnamese captions, totaling 19,250 captions.
To minimize inconsistencies in interpretation, strict annotation guidelines were established, inspired by the MS COCO dataset. These included: First, a minimum of ten Vietnamese words per caption. Second, captions should describe only visible objects and activities, excluding personal opinions, proper names, and numbers. Last, sentences should be written in continuous tense, with familiar English terms (e.g., "tennis") allowed for clarity.

The Indonesian dataset used in the study by \cite{mulyanto2019automatic} is FEEH-ID, which contains 8,099 images, each paired with five captions in Indonesian. The images were sourced from Flickr. The first 6,000 images from the training set were selected for their experiments. The captions were generated using a combination of Google Translate and manual translation by a professional English-Indonesian translator. The total vocabulary size of the dataset varied depending on the frequency of objects appearing in the images.

The Myanmar image captions corpus, developed by \cite{pa2020automatic}, was built using a subset of the Flickr8K dataset, which initially contains 8,092 images, each with five English captions. Due to time constraints, 3,000 images were selected, and five Myanmar-language captions were created for each image, totaling 15,000 captions. The dataset was constructed using two approaches: First, English captions were translated into Myanmar using an attention-based neural machine translation model, achieving a multi-BLEU score rate 13.93. Second, native speakers directly described the images in Myanmar, resulting in a vocabulary size of 3,138 words, with the longest caption containing 21 words. The dataset was divided into 2,500 images for training, 300 for validation, and 200 for testing.

The study by \cite{muhammad2022bornon} utilized three key datasets to generate Bengali captions for images. Together, these three Bengali datasets provide a comprehensive and diverse collection of images and captions, enabling more accurate and contextually rich Bengali caption generation. The datasets are:

\begin {enumerate}[a)]
\item Flickr8K-BN dataset: Contains 8,091 images, each with five Bengali captions. It covers a wide range of topics, including people, landscapes, animals, and everyday objects. The captions were originally in English and later translated into Bengali.
\item BanglaLekha dataset: Comprises 9,154 images, each with two Bengali captions. It focuses on themes such as animals, birds, food, trees, and buildings. It features a smaller vocabulary size than other datasets due to fewer captions per image.
\item Bornon dataset: Contains 4,100 images, each with five Bengali captions, totaling 20,500 captions. It covers diverse topics, including animals, people, food, weather, and vehicles. The Images were sourced from a personal photography club, and 17 native Bengali speakers annotated captions.
\end {enumerate}


\begin{table}[htbp]

\centering
\caption{Publicly available datasets for image captioning, \\detailing dataset names, sizes, and number of captions.}
\begin{tabular}{ccccc}

\label{tab:dataset}\\
\hline
\textbf{Datasets}  & \textbf{Train} & \textbf{Validate} & \textbf{Test} & \textbf{Captions} \\ \hline

\textbf{Flickr8k}  & 6,000          & 1,000             & 1,000         & 5                 \\ 
\textbf{Flickr30k} & 29,783         & 1,000             & 1,000         & 5                 \\ 
\textbf{MS COCO}   & 113,287        & 5,000             & 5,000         & 5                 \\ \hline

\end{tabular}

\end{table}

%
%

\section {Evaluation Metrics}
Measuring the accuracy of a generated text in describing an image is done more effectively through direct human judgments. However, expanding human evaluation is difficult due to the high amount of nonreusable human effort required. The following subsections introduce the commonly used evaluation metrics in image captioning. 
Table \ref{tab:symbols} provides definitions of key symbols used in the evaluation metrics. These definitions help understand the mathematical formulations behind BLEU, METEOR, CIDEr, ROUGE, and SPICE.

\begin{table*}[ht]
    \centering
    \caption{Definitions of symbols in evaluation metrics.}
    \resizebox{\textwidth}{!}{\begin{tabular}{lll}
        \hline
        \textbf{Symbol} & \textbf{Definition} & \textbf{Used In} \\ 
        \hline
        $n$-gram & Sequence of $n$ consecutive words in a sentence & BLEU, ROUGE \\ 
        $P_n$ & Precision of $n$-gram matches between generated and reference captions & BLEU \\ 
        BP & Brevity penalty (penalizes overly short captions) & BLEU \\ 
        $R$ & Recall – fraction of reference words covered by the candidate caption & METEOR, ROUGE \\ 
        $P$ & Precision – fraction of candidate words appearing in the reference caption & METEOR, ROUGE \\ 
        $F_1$ & Harmonic mean of precision and recall: $F_1 = \frac{2 P R}{P + R}$ & METEOR, ROUGE \\ 
        weight$(n)$ & Weight assigned to $n$-gram matches & BLEU \\ 
        geometric\_mean & Geometric mean of $n$-gram precision scores & BLEU \\ 
        $W$ & Set of words in candidate caption & CIDEr \\ 
        $W_r$ & Set of words in reference captions & CIDEr \\ 
        $freq(w, G)$ & Term frequency of word $w$ in $G$ (entire corpus) & CIDEr \\ 
        IDF$(w)$ & Inverse Document Frequency of $w$ & CIDEr \\ 
        $SPICE(S, R)$ & Graph-based semantic similarity between $S$ (candidate) and $R$ (reference) & SPICE \\ 
        \hline
    \end{tabular}}
    \label{tab:symbols}
\end{table*}

\subsection {Bilingual Evaluation Understudy BLEU}
Bilingual Evaluation Understudy (BLEU) is a metric used to evaluate the quality of machine-generated text. Assess individual text segments by comparing them to reference texts. The BLEU score varies depending on the number of reference translations and the length of the text produced. Generally, short-generated texts have higher BLEU scores ranging from 0 to 1. BLEU-1 uses unigram comparisons between candidate and reference sentences, while bigram comparisons are used for BLEU-2. An empirical maximum order of four optimizes correlation with human judgments. Unigram scores determine the adequacy of the BLEU metrics, while higher n-gram scores determine fluency \cite{papineni2002bleu}. The BLEU formula is defined as

\begin{equation}
\label{eq9}
 B L E U=B P \times \mathrm{exp}\left(\sum_{n=1}^N w_n \log \left(p_n\right)\right).
\end{equation}

The shortness penalty (BP) allows us to choose the candidate translation most similar to the reference translation in terms of length, word choice, and word order. It is calculated using an exponential decay given as

\begin{equation}
\label{eq10}
B P= \begin{cases}1 & m_c>m_r \\ \mathrm{e}^{\left(1-m_r / m_c\right)} & m_c \leqslant m_r\end{cases}
\end{equation}

The sum of the counts of the clipped $n$ gram of candidate sentences in corpus $CC_n$ is divided by the total number of candidate $n$-grams. $C_n$ is used to calculate the modified precision for each $n$-gram. It enables us to determine the sufficiency and fluency of the candidate translation relative to the reference translation as 

\begin{equation}
\label{eq11}
p_n=\frac{\sum_{C \in\{\text {Candidates}\}} \sum_{n-\text {gram}\in C} C C_N}{\sum_{C^{\prime} \in[\text {Candidates}\}} \sum_{n-\text {gram} \in C^{\prime}} C_N}.
\end{equation}

\subsection {The Recall Oriented Understudy for Gisting Evaluation ROUGE} 
The Recall-Oriented Understudy for Gisting Evaluation (ROUGE) \cite{lin2004rouge} is a set of metrics used to assess text summaries. It compares word sequences and word pairs with a reference database of human-written summaries in a given summary. The metric uses the longest common subsequence between a candidate sentence and a set of reference sentences to measure their similarity at the sentence level. ROUGE-1, ROUGE-2, ROUGE-W, and ROUGE-SU4 are different types of ROUGE used for various tasks, and the metric score ranges from 0 to 1.

Calculating the longest common subsequence (LCS), the longest matching sequence of words between the original and predicted summaries, forms the basis of the ROUGE algorithm. Unlike matching words consecutively, LCS allows for matches that reflect the word order at the sentence level. Additionally, LCS automatically includes common n-grams in sequence, removing the need to calculate predetermined $n$-gram sequences. Mathematically, ROUGE can be defined as

\begin{equation}
\label{eq12}
F_{l c s}=\frac{\left(1+\beta^2\right) \cdot R_{l c s} \cdot P_{l c s}}{R_{l c s}+\beta^2 \cdot P_{l c s}}.
\end{equation}

The LCS-based precision $P_{lcs}$ and the LCS-based recall $R_{lcs}$ can be calculated using the upper part of (\ref{eq13}) and (\ref{eq14}) for the sentence level, or can be calculated using the lower part of the same equations for the summary level.

\begin{equation}
\label{eq13}
P_{l c s}=\left\{\begin{array}{l}
\frac{l_{L C S}(X, Y)}{m_c} \\
\frac{\sum_{j=1}^{u_r} L C S \cup\left(r_j, c\right)}{m_c}
\end{array}\right.
\end{equation}

 \begin{equation}
 \label{eq14}
    R_{l c s}=\left\{\begin{array}{l}
   \frac{l_{L C S}(X, Y)}{m_r} \\
   \frac{\sum_{j=1}^{u_r} \operatorname{LCS} \cup\left (r_j, c\right)}{m_r}
   \end{array}\right.
 \end{equation}

where LCS is the longest common subsequence, $P_{lcs}$: LCS-based precision, $R_{lcs}$: LCS-based recall, $\beta$: $P_{lcs}/R_{lcs}$, $l_{LCS}$: length of the longest common subsequence of $X$ and $Y$, LCS $\boldsymbol{U}$ ($r_j$, $c$): LCS score of the union's longest common subsequence between a reference sentence and the candidate sentence.

\subsection {The Metric for Explicit Ordering Translation Evaluation METEOR}

The METEOR metric \cite{banerjee2005meteor} is designed to evaluate machine translation and is considered more valuable than the Blue metric. Its correlation with human evaluations is stronger. Meteor calculates a score by comparing a candidate sentence with a human-written reference sentence using generalized unigram matching. The score is computed based on the matched words' precision, recall, and alignment. When multiple reference sentences are involved, the candidate's final evaluation score is determined by choosing the best score among all independently computed ones.
METEOR incorporates stemming, synonym matching, and standard exact word matching, making it more effective at the sentence or segment level \cite{10.1145/3617592}. The maximum score can be estimated by computing the F-measure through explicit unigram matching (i.e., word-for-word matching) between the candidate and reference translations. The METEOR metric is defined as

\begin{equation}
\label{eq15}
METEOR=F_{\text {mean }} \cdot(1-p n)
\end{equation}\\

The chunks comprise adjacent unigrams in the reference and hypothesis to calculate the penalty $P_n$. The longer the adjacent mappings are between the candidate and the reference, the fewer chunks there are. The penalty is obtained by

\begin{equation}
\label{eq16}
p n=0.5 *\left(\frac{C h}{U_m}\right)^3
\end{equation}

A harmonic mean of precision and recall is determined as the F-mean, with a higher value on recall as

\begin{equation}
\label{eq17}
F_{\text {mean }}=\frac{10 . P . R}{R+9. P}
\end{equation}

and recall value $R$ as

\begin{equation}
\label{eq18}
R=\frac{M(c)}{U(r)}, 
\end{equation}

and precision $P$ as

\begin{equation}
\label{eq19}
P=\frac{M(c)}{U(c)}
\end{equation}

where $P_n$ is the penalty, $Ch$ is the number of chunks, $U_m$ is the number of unigrams that correspond between the candidate and the reference, $M(c)$ is the number of unigrams in the candidate sentence that are mapped, $U(r)$ is the total number of unigrams in the reference sentence, and $U(c)$ is the total number of unigrams in the candidate sentence.

\subsection {Consensus-based Image Description Evaluation CIDEr}

The Image Description Evaluation (IDE) tool uses the consensus-based Image Description Evaluation (CIDEr) metric to assess the similarity of a generated sentence to a set of human-authored ground truth sentences \cite{vedantam2015cider}. It employs a Term Frequency-Inverse Document Frequency (TF-IDF) weighting for each n-gram in the candidate phrase to encode their frequency in the reference sentences. This metric evaluates the grammar, relevance, and accuracy.

CIDEr was specifically designed to evaluate image captions and descriptions. Unlike other metrics that only work with five captions per image, it utilizes consensus through TF-IDF, making it unsuitable for analyzing the consensus between generated captions and human assessments \cite{10.1145/3617592}. Therefore, the average cosine similarity between the candidate and reference sentences is used to calculate the CIDEr score for $n$-grams of length $n$ as

\begin{equation}
\label{eq20}
\operatorname{CIDEr}_n\left(c, r_j\right)=\frac{1}{u_r} \sum_{j=1}^{u_r} \frac{g^n(c) \cdot g^n\left(r_j\right)}{\left\|g^n(c)\right\| \cdot\left\|g^n\left(r_j\right)\right\|}
\end{equation}

The weighting TF-IDF $g_k(r_j)$ for each $n$-gram $w_k$ of a reference sentence is defined as

\begin{equation}
\label{eq21}
g_k\left(r_j\right)=\frac{h_k\left(r_j\right)}{\sum_{w_l \in \Omega} h_l\left(r_j\right)} \log \left(\frac{|I|}{\sum_{I_p \in I} \min \left(1, \sum_j h_k\left(r_j\right)\right)}\right)
\end{equation}

Similarly, for $g_k(c)$, the candidate sentence is replaced by $r_j$ with $c$.  CIDEr is computed by combining the scores from $n$-grams of varying lengths as

\begin{equation}
\label{eq22}
\operatorname{CIDEr}\left(c, r_j\right)=\sum_{n=1}^N w_n \cdot \operatorname{CIDEr}_n\left(c, r_j\right)
\end{equation}

where $g_n(c)$ is a vector formed by all $n$-grams of length $n$ of the candidate sentence, $\left\| g_n(c)\right\|$ is the magnitude of the vector $g_n(c)$, $g_n(r_j)$ is a vector formed by all $n$-grams of length $n$ of the set of reference sentences, $\left\|g_n(r_j)\right\|$ is the magnitude of the vectors $g_n(r_j)$, $g_k(r_j)$ is TF-IDF weighting for each $n$-gram $w_k$ of the set of reference sentences, $g_k(c)$ is TF-IDF weighting for each $n$-gram $w_k$ of the candidate sentence, $h_k(r_j)$ is the number of occurrences of an $n$-gram $w_k$ in a reference sentence, $h_k(c)$ is the number of occurrences of an $n$-gram $w_k$ in the candidate sentence, $\Omega$ is the vocabulary of all $n$-grams, and $I$ is the set of all images in the dataset.

\subsection{The Semantic Propositional Image Caption Evaluation SPICE}

SPICE (Semantic Propositional Image Caption Evaluation) was developed to evaluate image captioning using semantic scene graphs \cite{anderson2016spice}. It is considered to be more accurate than human judgments. The process involves extracting information about various items, properties, and their relationships from image descriptions \cite{hossain2019comprehensive}. The captions are converted into scene graphs via semantic parsing. The similarity score between the generated and ground-truth caption scene graphs is calculated using precision and recall F1 scores. 

Precision is determined by the matching tuples between the logical tuples for generated and reference captions divided by the total number of logical tuples in the generated caption set. For recall, the matching tuples are divided by the total number of logical tuples in the reference caption set. The F1 score (SPICE) is calculated using precision and recall \cite{zohourianshahzadi2021neural}. Scene graphs ($G(c)$ and $G(S_r)$ are created from candidate and reference captions, respectively), and the F score is calculated using the conjunction of logical tuples representing semantic propositions in the scene graph. SPICE can be calculated using the scene graphs of all reference sentences as 

\begin{equation}
\label{eq23}
\operatorname{SPICE}\left(c, S_r\right)=F_1\left(c, S_r\right)=\frac{2 \cdot P\left(c, S_r\right) \cdot R\left(c, S_r\right)}{P\left(c, S_r\right)+R\left(c, S_r\right)}\\
\end{equation}

Precision and recall are calculated as in (\ref{eq24}) and (\ref{eq25}), respectively.

\begin{equation}
\label{eq24}
P=\frac{\left|T(G(c)) \bigotimes T\left(G\left(S_r\right)\right)\right|}{|T(G(c))|}
\end{equation}

\begin{equation}
\label{eq25}
R=\frac{\left|T(G(c)) \bigotimes T\left(G\left(S_r\right)\right)\right|}{\left|T\left(G\left(S_r\right)\right)\right|}
\end{equation}

where $G(c)$ is the scene graph of the candidate sentence, $G(r_j)$ is the scene graph of each reference sentence, $G(S_r)$ is the scene graph of all reference sentences, $O(c)$ is set of objects in the candidate sentence, $E(c)$ is set of attributes in the candidate sentence, $K(c)$ is set of relations in the candidate sentence, $T$ is the function that allows us to return logical tuples.

\section {Limitation and Challenges}

The development of image captioning models faces several challenges, including exploding gradients and the generation of incorrect sentences. Most modern algorithms rely on Recurrent Neural Networks and Long-Short-Term Memory Networks, which can suffer from vanished gradients and require significant resources, making them less hardware-friendly \cite{pa2020automatic}. Although Generative Adversarial Networks \cite{zhang2019image} offer a promising alternative, they come with their own set of issues, such as the difficulty of training due to the discrete nature of GAN \cite{abu2022effect, abu2021paired}. Another approach involves using semantic feature vectors or focusing on object-region relationships \cite{gong2019enhanced} or focusing on object-region relationships \cite{biswas2020towards} \cite{pedersoli2017areas} \cite{yu2019multimodal}.

Current methods for evaluating caption quality use logarithmic likelihood scores and automated metrics such as BLEU, METEOR, ROUGE, CIDEr, and SPICE. However, these metrics often do not correlate well with human evaluations. Despite SPICE and CIDEr being closer to human judgment, they still challenge optimization \cite{hossain2019comprehensive}. These automated measures mainly focus on lexical or semantic data and do not fully capture the complex relationships between words and objects \cite{vaswani2017attention}. To enhance captioning models, it is essential to improve automatic evaluation methods to more accurately reflect human judgments and address gaps in understanding object and word relationships \cite{li2019entangled}.

As image captioning technology advances, several future challenges must be addressed to enhance its effectiveness. One major challenge is improving the adaptability of captioning systems to diverse languages and cultural contexts. Current models often struggle with non-English languages and culturally nuanced images, leading to less accurate or contextually relevant captions. Another challenge is the need for more comprehensive and inclusive datasets. Many existing datasets are limited in scope or do not represent a wide range of cultural and contextual variations, affecting the generalizability of captioning systems. Additionally, there is a growing need to develop models that can understand and generate captions for complex, abstract, or ambiguous images, where traditional methods may fall short. Ensuring these systems can operate effectively in real-world scenarios, including understanding user-specific contexts and preferences, is also crucial.

\section{Conclusion and Research Opportunities}

Image captioning has emerged as a crucial task at the intersection of computer vision and natural language processing, enabling machines to understand and describe visual content. This survey has comprehensively reviewed attention-based transformer models, covering their architectures, evaluation metrics, datasets, and multilingual applications. We highlighted the transition from traditional template-based approaches to deep learning-driven transformer models, emphasizing the role of attention mechanisms in improving caption quality.
Despite significant advancements, key challenges include handling complex scene compositions, improving caption fluency in low-resource languages, and ensuring factual accuracy in generated descriptions. Future research opportunities could focus on:

\begin{enumerate}[a)]
\item Multimodal learning and cross-domain image captioning: Integrating vision, language, and other sensory inputs for richer, more context-aware captions.

\item Multilingual and cross-lingual captioning: Expanding datasets and improving transfer learning techniques for non-English languages.

\item Real-time and interactive captioning: Optimizing models for assistive AI, augmented reality, and robotics applications.

\item Applications in novel domains: Extending image captioning to forensic analysis, cultural heritage, and personalized AI assistants.

\end {enumerate}

As AI evolves, attention-based transformer models will remain superior in bridging the gap between vision and language. Addressing the challenges and opportunities outlined in this survey will be essential for realizing the full potential of image captioning in diverse real-world applications.

\subsubsection*{Abbreviations}
\noindent The abbreviations used in this manuscript are given in Table \ref{tab:abbreviations}.

\begin{table}[H]
    \centering
    \caption{List of abbreviations}
      \resizebox{9cm}{!}{\begin{tabular}{ll}
        \hline
        \textbf{Abbreviation} & \textbf{Full Form} \\ 
        \hline
        AI & Artificial Intelligence \\ 
        CNN & Convolutional Neural Network \\ 
        RNN & Recurrent Neural Network \\ 
        LSTM & Long Short-Term Memory \\ 
        GRU & Gated Recurrent Unit \\ 
        NLP & Natural Language Processing \\ 
        CV & Computer Vision \\ 
        GAN & Generative Adversarial Network \\ 
        BLEU & Bilingual Evaluation Understudy (Metric) \\ 
        METEOR & Metric for Evaluation of Translation with Explicit ORdering \\ 
        CIDEr & Consensus-based Image Description Evaluation \\ 
        ROUGE & Recall-Oriented Understudy for Gisting Evaluation \\ 
        SPICE & Semantic Propositional Image Caption Evaluation \\ 
        VLP & Vision-Language Pre-training \\ 
        MS COCO & Microsoft Common Objects in Context (Dataset) \\ 
        SLR & Systematic Literature Review \\ 
        \hline
    \end{tabular}}
    \label{tab:abbreviations}
\end{table}


\begin{table*}[htbp] 
    \centering
    \caption{Summary of multilingual image captioning models}
    \label{tab:Multilingual image captioning models}
    \footnotesize
\begin{tabularx}{\textwidth}{ p{1.8cm} p{1.5cm} p{1.5cm} p{2.5cm} p{1.5cm} p{12cm}}   
    \hline

\clearpage
\textbf{Reference} &
\textbf{Image model} &
\textbf{Language model} &
\textbf{Dataset} &
\textbf{Dataset's language} &
\textbf{Improvement} \\ \hline

Zhang 2021\\\cite{zhang2021consensus} &
  LSTM &
  LSTM &
  Flickr30k &
  English &
 Provided fine-grained information among objects \\
  \hline

Al 2018 \\\cite{al2018automatic} &
  VGG &
  RNN-LSTM & 
  MSCOCO, Flickr8k &
  Arabic &
  The model can achieve excellent results with larger corpus\\
  \hline
  
Chen 2019 \\\cite{chen2019news} &
  CNN &
  RNN &
  Daily Mail &
  English &
  Considering both the news; image and text \\
  \hline
  
Biswas 2020\\\cite{biswas2020towards} &
  CNN &
  LSTM &
  MSCOCO &
  English &
 Provided further improvements in image captioning \\
\hline
  
Mualla 2018\\ \cite{mualla2018development} &
  CNN &
  LSTM &
 Flickr8k &
  Arabic &
  The English-based model had the best performance \\
  \hline
  
Eljundi 2020\\\cite{eljundi2020resources} &
VGG16 &
LSTM layer &
Flickr8K &
Arabic &
Showed the superiority of the end-to-end model \\
  \hline
  
Do 2020\\\cite{do2020reference} &
  CNN &
  GRU &
  Flickr30K MSCOCO &
  English &
  PoS gave the best performance\\
  \hline
  
Saleh 2019\\\cite{saleh2019towards} &
  CNN &
  LSTM &
  MSCOCO &
  Arabic &  Convert stories to images support the meaning\\
  \hline

Cheikh 2020\\\cite{cheikh2020active} &
  CNN &
  LSTM &
  ArabicFlickr1K &
  Arabic &
  Applied human annotators \\
\hline

Mulyanto 2019\\\cite{mulyanto2019automatic} &
  CNN &
  LSTM &
  Flickr &
  Indonesian & The test set provided promising results \\
\hline

Tien 2020\\\cite{tien2020image} &
  CNN &
  bi-RNN &
  MSCOCO  &
  Vietnamese &
  A solution to the problem of unknown words\\
\hline
  
Pa 2020\\\cite{pa2020automatic} &
  CNN &
  LSTM &
  Flickr8k&
  Myanmar &
  Automatic translation reduced the manual captioning time \\
\hline
  
Yu 2019\\\cite{yu2019multimodal}&
  R-CNN &
  LSTM layer &
  MSCOCO &
  English & Three types of relations\\
  \hline
  
He 2020\\\cite{he2020image}&
 Transformer &
  Transformer &
  MSCOCO &
  English &
 Query region: parent, neighbor, or child \\
  \hline
  
Pedersoli 2017\\\cite{pedersoli2017areas}&
 CNN &
  GRU &
  MSCOCO &
  English &
  Modeled the direct dependencies between words and image \\
  \hline
  
  
Fei 2022\\\cite{fei2022attention} &
  Transformer &
  Transformer &
  MSCOCO & English &  Image region features to direct attention alignment \\
\hline
 
WEI 2022\\\cite{WEI2022104574} &
 RNN &
  Transformer &
  MSCOCO &
  English &
  Capture the dependencies within the image areas and regions \\
\hline
  
Wang 2022\\\cite{wang2022geometry} &
   LSTM &
  Transformer &
  MSCOCO Flickr30k &
  English &
  Utilizes geometry relations \\
\hline
  
Yan 2022\\\cite{9381876} &
  FR-CNN &
  Transformer &
  MSCOCO &
  English &
  Reduces the misinformation \\
\hline
  
Lu 2023\\\cite{lu2023full} &
  Transformer &
  Transformer &
  MSCOCO &
  English &
  Common self-attentive mechanism \\
\hline
  
Dubey 2023\\\cite{dubey2023label} &
  FRCNN &
  Transformer &
  MSCOCO &
  English &
 Relate objects based on localized ratios \\
  \hline

Wang 2021\\\cite{wang2021reasoning} &
  FRCNN &
  Transformer &
  MSCOCO &
  English &
 Dynamic attention aggregation \\
\hline
  
Wang 2020\\\cite{wang2020transformer} &
 R-CNN &
  Transformer &
  MSCOCO &
  English &   Capture critical objects and relationships \\
  \hline
  
Guo 2020\\\cite{guo2020normalized} &
  FRCNN &
  Transformer &
  MSCOCO &
  English &  Provided a unique normalization method \\
  \hline
  
Jiang 2021\\\cite{jiang2021multi} &
  FR-CNN &
  LSTM &
  MSCOCO &
  English &  Simplify the model and increase efficiency \\
  \hline
  

Liu 2022\\\cite{liu2022image} & 
  Transformer &
  Transformer &
  MSCOCO &
  English &
  Visual attention and keyword attention \\
\hline

Liu 2022\\\cite{liu2022remote} &
  ResNet50 &
  Transformer &
  RSICD &
  English &
  Effectively construct sentences \\
  \hline
  
Kandala 2022\\\cite{kandala2022exploring} &
  Transformer &
  Transformer &
 UC-Merced &
  English &
  Self-attention improve the performance \\
  \hline
  
Nguyen 2022\\\cite{nguyen2022grit}  &
  Transformer &
  Transformer &
  MSCOCO &
  English &
  Transformer get beyond the drawback of CNN \\
  \hline
  
Tan 2022\\\cite{tan2022acort} &
  Transformer &
  Transformer &
  MSCOCO &
  English &  Fewer parameters without loss of performance\\ \hline
  
Kumar 2022\\\cite{kumar2022dual}&
  Inception-V3 &
  Transformer &
  Flickr MSCOCO, &
  English &  intra- and inter-model interactions \\
\hline
  
Wang 2022\\\cite{wang2022dm} &
  Transformer &
  Transformer &
  MSCOCO &
  English &  high-order intra- and inter-feature interactions\\
  \hline
  
Ma 2023\\\cite{ma2023towards} &
  Transformer &
  Transformer &
  Flickr MSCOCO &
  English & 
  Local visual modeling with grid features\\ 
  
  \hline

 \end{tabularx} %
\end{table*}
\clearpage




\appendix


 \bibliographystyle{elsarticle-num} 
 \bibliography{cas-refs}





\end{document}